\documentclass[letterpaper,twocolumn,10pt]{article}
\usepackage{usenix-2020-09}

\usepackage{tikz}
\usepackage{amsmath}

\usepackage{custom_macros} %
\usepackage{dsfont}
\usepackage{enumitem}
\usepackage{booktabs}
\usepackage{amssymb}
\usepackage{amsthm}
\usepackage{float}
\usepackage[nobiblatex]{xurl} %
\usepackage{adjustbox}
\usepackage{multirow}

\renewcommand{\footnoterule}{%
  \kern -3pt
  \hrule width 1in
  \kern 2pt
}

\let\OLDthebibliography\thebibliography
\renewcommand\thebibliography[1]{
  \OLDthebibliography{#1}
  \setlength{\parskip}{2pt}
  \setlength{\itemsep}{0pt plus 0.3ex}
}

\usepackage[small,bf]{caption}

\newif\ifusenixfinal
\usenixfinalfalse

\begin{document}
\theoremstyle{definition}
\newtheorem{defn}{Definition}[section]

\date{}

\renewcommand*{\thefootnote}{\fnsymbol{footnote}}
\title{\Large \bf A Linear Reconstruction Approach for Attribute Inference Attacks against Synthetic Data\ifusenixfinal \thanks{An extended version of this paper can be found \href{https://arxiv.org/abs/2301.10053}{here}.} \else \thanks{This is an extended version of our paper that appeared at USENIX Security 2024.} \fi}

\renewcommand*{\thefootnote}{\arabic{footnote}}
\setcounter{footnote}{0}

\author{
    {\rm Meenatchi Sundaram Muthu Selva Annamalai} \\
    University College London
    \and
    {\rm Andrea Gadotti} \\
    University of Oxford
    \and
    {\rm Luc Rocher} \\
    University of Oxford
}

\maketitle

\ifusenixfinal \pagenumbering{gobble} \else \pagenumbering{arabic} \fi

\begin{abstract}
Recent advances in synthetic data generation (SDG) have been hailed as a solution to the difficult problem of sharing sensitive data while protecting privacy. SDG aims to learn statistical properties of real data in order to generate ``artificial'' data that are structurally and statistically similar to sensitive data. However, prior research suggests that inference attacks on synthetic data can undermine privacy, but only for specific outlier records.

In this work, we introduce a new attribute inference attack against synthetic data. The attack is based on linear reconstruction methods for aggregate statistics, which target all records in the dataset, not only outliers. We evaluate our attack on state-of-the-art SDG algorithms, including Probabilistic Graphical Models, Generative Adversarial Networks, and recent differentially private SDG mechanisms. By defining a formal privacy game, we show that our attack can be highly accurate even on arbitrary records, and that this is the result of individual information leakage (as opposed to population-level inference).

We then systematically evaluate the tradeoff between protecting privacy and preserving statistical utility. Our findings suggest that current SDG methods cannot consistently provide sufficient privacy protection against inference attacks while retaining reasonable utility. The best method evaluated, a differentially private SDG mechanism, can provide both protection against inference attacks and reasonable utility, but only in very specific settings. Lastly, we show that releasing a larger number of synthetic records can improve utility but at the cost of making attacks far more effective.
\end{abstract}
\section{Introduction}
\label{sec:intro}
Modern scientific research benefits from large datasets of personal data, facilitated by data sharing between organizations. In the past decade, data breaches, privacy threats, and lost public trust have challenged access to data, putting reproducibility and open data efforts at risk. Data anonymization---the process of sanitizing data so that individuals are no longer identifiable---is one of the main ways to share data while minimizing privacy risks for data subjects. However, a large body of research has shown that traditional anonymization techniques are generally inadequate and do not provide a good tradeoff between privacy and statistical utility for modern data~\cite{ohm2009broken, rocher2019estimating}.

Synthetic data generation (SDG) has emerged as a potential solution to provide both high utility and privacy protection. This approach relies on algorithms that learn the underlying distribution of the data, to then generate structurally and statistically similar data. The resulting synthetic data may not contain any original record, yet can be seamlessly integrated into analysis pipelines at little to no cost. One of the main use cases of synthetic data is in tabular data publishing, with the US National Institute of Standards and Technology (NIST) suggesting that synthetic data can ``serve as a practical replacement of the original sensitive data''~\cite{dpsynth2018}. The UK Royal Society further argued that it ``retain(s) the statistical properties of the original dataset'' such that ``the anonymity of the original dataset is not compromised''~\cite{royalsocietysynth}.

Despite these positive early evaluations, the fact that synthetic records are ``artificial'' does not, per se, guarantee that the privacy of all individuals in the original dataset are protected. In the past few years, the privacy guarantees provided by generative models and synthetic data have come under heightened scrutiny~\cite{carlini2021extracting, stadler2022synthetic, yale2019assessing, oprisanu2021utility, chen2020gan, hilprecht2019monte, hayes2017logan, giomi2022unified, houssiau2022tapas}. For instance, having full access to the internal parameters of the trained generative model allows adversaries to run powerful attacks against high-dimensional synthetic data~\cite{hayes2017logan, carlini2021extracting, hilprecht2019monte, chen2020gan}. However, this adversarial model, referred to as the \textit{white-box} setting, is often unrealistic as the trained model is typically not shared in the context of synthetic data~\cite{zhang2022membership, royalsocietysynth}.

In the \textit{black-box}\footnote{We note that black-box is used also with a slightly different meaning in the context of attacks against machine learning models~\cite{nasr2019comprehensive}. However, in the context of synthetic data, this taxonomy~\cite{houssiau2022tapas} is more suited as it reflects the way synthetic data is generated and released in practice.} setting, where the adversary has exact knowledge on the specifications of the SDG algorithm along with some auxiliary knowledge, Stadler et al.~\cite{stadler2022synthetic} propose a membership inference attack (with an adversary attempting to predict presence in the training dataset), adapting the shadow modelling technique from Shokri et al.~\cite{shokri2017membership}. They show that information leakage can in fact be exploited without directly relying on overt memorization by the generative model, but only for specific outlier records.

In the most restrictive scenario where the adversary has no knowledge of the SDG algorithm (\textit{no-box} setting), while prior work has proposed attacks, they have limited effectiveness in practice. Against generative models that memorize and output training data, Chen et al.~suggested a simple heuristic attack---searching for the closest synthetic record to a victim's record~\cite{chen2020gan}. However, their attack has limited effectiveness in the context of synthetic tabular data~\cite{houssiau2022tapas}. Stadler et al.~proposed a second attack that uses machine learning to infer the secret attribute of a target record (attribute inference attack). However, in their experiments, the adversary can precisely infer the target record's sensitive attribute even when the target record is \textit{not present in the original dataset}. This means that the attack may, in principle, be exploiting population-level associations, making it difficult to assess whether synthetic data actually ``remembers'' information about the original records (nor, arguably, on whether the SDG mechanism protects privacy~\cite{bun2021statistical}). This issue has been referred to as the base rate problem by Houssiau et al.~\cite{houssiau2022tapas}.

\paragraph{Contributions.}
Here, we study the privacy leakage from synthetic tabular data in the \textit{no-box} setting. To that end, we introduce a new attribute inference game with a \emph{partially informed} adversary who has access to two datasets: the synthetic dataset and the ‘quasi-identifiers’ of the records from the original dataset (i.e. all the attributes except the one that is to be inferred). Under our privacy game, the adversary tries to infer the \textit{randomized} secret attribute of a given record. This solves the base rate problem mentioned above and enables us to evaluate information leakage about individual users. Our main contributions are as follows:
\begin{enumerate}
    \item We propose a new attribute inference attack on synthetic data that leverages overly-accurate aggregate statistics computed on the synthetic data. Specifically, our attack does not directly rely on overfitting or overt memorization by the generative model. Instead, the attack is based on linear reconstruction attacks~\cite{kasiviswanathan2013power,cohen2018linear}, which aims to reconstruct the whole dataset (or column) without targeting specific users or outliers.
    \item We show that many users can in fact be vulnerable to attribute inference attacks against synthetic data, not just outliers. Specifically, our attack can infer a target record's secret attribute with up to 94.8\% probability, compared to a random baseline of 50\%. Our attack can increase accuracy by up to 9.6 percentage points compared to prior attacks, when evaluated on our privacy game. This shows that simple heuristics, such as the distance to closest synthetic record, can underestimate the privacy risks of synthetic data.
    \item We provide a quantitative framework to evaluate privacy-utility tradeoffs of synthetic data. In particular, we show that, \emph{for the same trained SDG model}, releasing a larger number of synthetic records can result in better utility but worse privacy. To the best of our knowledge, the impact of synthetic data size has not been studied for privacy, and has been mentioned only marginally for utility~\cite{aydore2021differentially}. This has important practical implications, suggesting that empirical evaluations on small benchmark datasets can strongly underestimate privacy risk if larger synthetic datasets are then released.
    \item We find that synthetic data, on its own, can hardly provide both good utility and good privacy protections---not only for outliers but even for arbitrary records. As an example, in our empirical evaluations, none of the state-of-the-art SDG algorithms could provide both accurate observations with a mean relative error below 0.20 and attack accuracy below 60\%.
    \item Lastly, we investigate the effect of formal differentially private (DP) defenses on synthetic data generation. We show that, in specific cases, recent DP mechanisms can offer good privacy-utility tradeoffs in practice, at the cost of weak formal guarantees.
\end{enumerate}
\section{Related Work}
\label{sec:related_work}
Since the first membership inference attack (MIA) against machine learning models was proposed by Shokri et al.~\cite{shokri2017membership}, a long line of work has investigated MIAs against generative models. The models under study include large language models~\cite{carlini2021extracting}, generative adversarial networks~\cite{hayes2017logan, chen2020gan, hilprecht2019monte}, diffusion models~\cite{zhu2023data}, and variational autoencoders~\cite{hilprecht2019monte}. These prior work have focused both on the \textit{white-box} setting (where an adversary has full access to the trained model) and on the \textit{black-box} setting (where an adversary has exact knowledge of the specifications of the generative model). However, in the context of synthetic data, the underlying generative model is typically not shared and the adversary only has access to the generated synthetic dataset~\cite{zhang2022membership, royalsocietysynth}. Therefore, these prior attacks propose strong adversarial models that may be unrealistic in practice. Here, we instead focus on the \textit{no-box} setting: the adversary has access to the anonymised synthetic dataset but no information about the underlying generative model or even the specifications of the synthetic data generation algorithm~\cite{houssiau2022tapas}.

Early work on attacks against synthetic data have proposed simple heuristics, with the intuition that synthetic datasets might violate privacy if they overtly memorize and output the training data. In such cases, records in the original dataset would appear as-is or with small differences in the synthetic dataset. To quantify privacy leakages, researchers have suggested to compute the distance to closest synthetic record~\cite{lu2019empirical, hilprecht2019monte}, which we refer to as the DCR heuristic. By introducing artificial privacy leakage, Giomi et al.~suggest that the DCR heuristic works best when the generative model overtly memorizes training data by generating synthetic records that are very close to or almost the same as records in the training data~\cite{giomi2022unified}. However, information can leak in many more complex ways than direct memorization as has been previously shown for other types of analyses such as aggregate statistics~\cite{pyrgelis2018knock,pyrgelis2017what,dick2022confidence} and machine learning~\cite{hayes2017logan,ye2022enhanced,carlini2022membership}.

In the context of synthetic data, Stadler et al.~\cite{stadler2022synthetic} were the first to show that individual-level information can in fact be leaked from synthetic data beyond simple heuristics. In their membership inference attack, Stadler et al.~follow the shadow modelling approach of Shokri et al.~and show that 4 out of 5 outlier records tested were vulnerable to membership inference attacks. However, even amongst the outlier records, 1 out of 5 achieved close to perfect privacy protections. Additionally, under their framework, all 5 randomly chosen records achieve close to perfect privacy protections which suggests that there is little to no privacy leakage for the average user from synthetic data. While attacks against a small minority of users can be useful to measure theoretical risks, they may not be necessarily relevant in practice especially if the adversary does not have a precise way to recognize vulnerable users~\cite{cohen2022attacks,gadotti2022pool}.

On the other hand, attribute inference attacks are an understudied area of research in the context of synthetic data. Stadler et al.~additionally propose an attribute inference attack where the adversary trains a machine learning classifier on the synthetic data to predict the secret attribute. The adversary then uses this classifier to infer the secret attribute for the target record. However under their privacy game, the adversary could infer the target record's sensitive attribute \textit{even if the target record was not in the original dataset} thus making it hard to determine if enough information pertaining to a specific target record was present in the synthetic data. Therefore, whether synthetic data are vulnerable to attribute inference attacks remains an open question.
\section{Preliminaries}
\label{sec:prelim}
\paragraph{Datasets.}
A \emph{dataset}, $\mathcal{D}$, is a multiset of $n$ records from a domain with $d$ discrete attributes $\mathcal{X} = \mathcal{X}_1 \times ... \times \mathcal{X}_d$. For simplicity, we assume that the domain of each attribute, $\mathcal{X}_i$, is finite so that each attribute is discrete\footnote{In the context of synthetic data, this assumption is often true in practice, as continuous attributes are often discretized before the generative model is trained \cite{mckenna2022aim, aydore2021differentially, zhang2017privbayes}.}. $\mathcal{D}$ can be seen as a matrix of $n$ records and $d$ attributes, and we write $\mathcal{D}$ as $\mathcal{D} = \{X | \mathbf{y} \}$, where $X$ is the first $d-1$ columns and $\mathbf{y}$ is the last column. In the next section, we will consider, without loss of generality, the $d$-th attribute as the secret that the adversary tries to infer.

Following prior work in the area of linear reconstruction attacks \cite{dinur2003revealing, cohen2018linear, dwork2007price}, we assume for simplicity that the secret attribute is binary i.e. $\mathcal{X}_d = \{0, 1\}$ and $\mathbf{y} \in \{0, 1\}^n$.

\paragraph{Synthetic data generation.}
A \emph{synthetic data generation (SDG) algorithm} is any algorithm (deterministic or randomized) that takes as input an \emph{original dataset} $\mathcal{D}$ and outputs a \emph{synthetic dataset} $\mathcal{S} \sim \text{SDG}(\mathcal{D}, m)$. We denote by $m$ the number of generated records, called \emph{synthetic data size}. We consider only SDG algorithms where original ($\mathcal{D}$) and synthetic ($\mathcal{S}$) records are from the same domain $\mathcal{X}$.

In practice, most SDG algorithms work by training a generative model $\mathcal{G}$ on the original dataset $\mathcal{D}$. The trained model $\mathcal{G}(\mathcal{D})$ can be modelled as a stochastic function that generates synthetic data records. As we are not concerned with attacking the underlying generative model in this work, we abstract out these details and consider any SDG algorithm as a function that outputs records from the domain $\mathcal{X}$.

\paragraph{Marginal queries.}
Marginal queries, also called cross tabulations, are one of the most important and useful statistics computed on tabular datasets \cite{dpsynth2018}. Informally, they take the form ``what fraction of people in the dataset are \textit{married}, \textit{employed} and \textit{have a college degree}?” and encode associations in the dataset. Formally, a $k$-way marginal query is defined as follows:
\begin{defn}[\textbf{$k$-way marginal query}]
Given a dataset $\mathcal{D}$ over a data domain with $d$ attributes, $\mathcal{X} = \mathcal{X}_1 \times ... \mathcal{X}_d$, a $k$-way marginal query is defined by a subset of attributes $\Attr \subseteq \{1, 2, ..., d\}$, $|\Attr| = k$ and corresponding values for each of the attributes $v \in \prod_{i \in \Attr} \mathcal{X}_i$. Given the pair $(\Attr, v)$, define $\mathcal{X}(\Attr, v) = \{x \in \mathcal{X} : x_i = v_i\; \forall i \in \Attr \}$. The corresponding $k$-way marginal query is then defined as follows where $\mathds{1}$ is the indicator function that maps elements of the set to 1 and the rest to 0:  
\begin{equation*}
    Q_{\Attr, v}(\mathcal{D}) = \frac{1}{|\mathcal{D}|} \sum_{x \in \mathcal{D}} \mathds{1} (x \in \mathcal{X}(\Attr, v))
\end{equation*}
\end{defn}
\section{Attribute Inference Attacks}
\label{sec:attack}

Successful attribute inference attacks have been considered as an important privacy risk in machine learning~\cite{fredrikson2014privacy} and data publishing~\cite{komarova2015estimation}. However, in recent years, the validity of some attribute inference attacks in the literature has come into question~\cite{jayaraman2022attribute,houssiau2022tapas,bun2021statistical}. In order to mitigate these concerns, we first formalize attribute inference in a new privacy \emph{game}. Thereafter, we introduce a new attribute inference \emph{attack} against synthetic data under our privacy game.

\subsection{Attribute Inference Privacy Game}
\label{subsubsec:ai_game}
\paragraph{Threat model.} Before formalizing the attribute inference attack model as a game, we provide an intuitive description. We consider an original dataset $\mathcal{D} = \{X|\mathbf{y}\}$ and a target user $u$ that is known to belong to the original dataset. We assume that the adversary has access to the target's partial record $\mathbf{x}_u \in X$, and aims to infer the value of the target's secret attribute $\mathbf{y}_u$. Our threat model follows the \emph{no-box} setting~\cite{houssiau2022tapas}: the adversary has access only to the synthetic dataset $\mathcal{S}$ and no other information on the synthetic data generation procedure.
 
While previous work~\cite{stadler2022synthetic,chen2020gan,hayes2017logan} often assumes that the adversary has access to a reference dataset (typically sampled from the same distribution as the target dataset), we do not make such assumption. Instead, we assume that the adversary is \emph{partially informed}: the adversary has access to $X$ (the adversary's auxiliary information), but not to the secret attribute values $\mathbf{y}$. We call $X$ the matrix of \emph{quasi-identifiers}, representing information that may be publicly available through additional data releases. Table~\ref{table:notation} provides a summary of the notation and the adversary's knowledge. We motivate the use of this threat model in Section~\ref{sec:discussion-partially-informed}.

In this setting, an SDG algorithm protects against attribute inference if the synthetic dataset $\mathcal{S}$ does not leak any information about the secret attribute values $\mathbf{y}$.

We now provide the formal description of the game:

\vspace{10pt}
\noindent\fbox{%
\parbox{.97\linewidth}{%
\minipage[t]{\dimexpr\linewidth-2\fboxsep-2\fboxrule\relax}
    \textbf{Attribute Inference Privacy Game.}
    \begin{itemize}
        \item \emph{Step 1.} Challenger samples original dataset $\mathcal{D} = \{X | \mathbf{y}\}$ with $n$ records.
        \item \emph{Step 2.} Challenger selects a random target record $u$ such that its quasi-identifiers $\mathbf{x}_u$ are unique in $X$.
        \item \emph{Step 3.} The Challenger then randomizes its secret attribute to create the target dataset. That is, $\mathcal{D}' = \{X | \mathbf{y}'\}$ such that $\mathbf{y}'_i = \mathbf{y}_i\; \forall i \in [n], i \neq u$ and $\mathbf{y}'_u \leftarrow \{0, 1\}$ uniformly at random.
        \item \emph{Step 4.} Challenger generates synthetic dataset $\mathcal{S}$ with $m$ records based on the target dataset $\mathcal{D}'$. That is, $\mathcal{S} \sim \SDG(\mathcal{D}', m)$.
        \item \emph{Step 5.} Challenger sends the synthetic dataset $\mathcal{S}$, the quasi-identifiers of the target record $\mathbf{x}_u$, and the quasi-identifiers of all records $X$ to Adversary.
        \item \emph{Step 6.} Adversary runs an attack $\mathcal{A}$ to guess the secret bit of the user $\widehat{\mathbf{y}'_u} = \mathcal{A}(\mathcal{S}, \mathbf{x}_u; X)$
    \end{itemize}
    \centering{\textbf{Adversary wins game if $\widehat{\mathbf{y}'_u} = \mathbf{y}'_u$}}
\endminipage}
}
\\

In Step 2, we ensure that the target's quasi-identifiers are unique. This is a common assumption in the literature~\cite{cretu2022querysnout} and simplifies the evaluation of the attack. When two or more records share the same quasi-identifiers, it might not be possible to infer their correct secret attribute even with access to the target dataset $\mathcal{D}'$, unless they also share the same secret attribute. Furthermore, this reflects the typical re-identification scenario where a user can be uniquely identified from their quasi-identifiers~\cite{sweeney1997weaving,lu2019empirical,gadotti2019signal} and has been shown to be realistic in high dimensional settings~\cite{rocher2019estimating}.

In Step 3, we randomize the secret attribute of the target record $u$ in order to measure individual-level leakage as opposed to population-level inferences. In Section~\ref{sec:discussion-privacy-game}, we discuss this aspect and compare our Attribute Inference Privacy Game with prior games from the literature.

\begin{table}[htb]
    \centering
    \scriptsize
    \begin{tabular}{r l l}
        \toprule
        {\textbf Symbol} & {\bf Description} & {\bf Known to $\Adv$} \\ 
        \midrule
        $\mathcal{D} = \{X|\mathbf{y}\}$ & Original dataset & \No \\
        $\mathcal{S}$ & Synthetic dataset & \Yes \\
        $\SDG$ & Synthetic data generation function & \No \\
        $n$ & Number of records in original dataset & \Yes \\
        $m$ & Number of records in synthetic dataset & \Yes \\
        $X$ & Quasi-identifiers in original \& target dataset & \Yes \\
        $\mathcal{X}$ & Combined domain of attributes in target dataset & \Yes \\
        $\mathcal{X}_i$ & Domain of $i$\textsuperscript{th} attribute in target dataset & \Yes \\
        $\mathbf{y}$ & Secret attributes of records in original dataset & \No \\
        \midrule
        $\Adv$ & Adversary & \\
        $\mathcal{D}' = \{X|\mathbf{y}'\}$ & Dataset targeted by adversary in game & \No \\
        $\mathbf{y}'$ & Secret attributes of records in target dataset & \No \\
        $\mathbf{x}_u$ & Quasi-identifiers of target record & \Yes \\
        $\mathbf{y}'_u$ & Secret attribute of target record in target dataset & \No \\
        $\widehat{\mathbf{y}'_u}$ & Adversary's guess for the target's secret attribute & \Yes \\
        \bottomrule
    \end{tabular}
    \vspace{0.5cm}
    \caption{Main notations for datasets, adversary, and game.}
    \label{table:notation}
\end{table}

\subsection{Linear reconstruction attacks against synthetic data}
\label{subsec:advs}
In this work, we introduce a new attribute inference attack under the Attribute Inference Privacy Game presented above. Our attack is based on linear reconstruction attacks, that we extended and adapt to specifically perform attribute inference on synthetic data. To the best of our knowledge, linear reconstruction attacks have not been previously studied in the context of synthetic data.

Synthetic data aims to serve as a safe replacement for original data while retaining the statistical properties of the original dataset. Therefore, we should expect that statistical queries on the synthetic data are reasonably accurate, with low error compared to original data. Intuitively, this is what our attack exploits. Given the synthetic dataset and the quasi-identifiers of the records in the target dataset, the adversary generates a set of statistical queries that should be preserved by the synthetic data. The adversary then estimates the most likely vector of secret attributes by minimising the error between these statistical queries on synthetic and reconstructed data. Hence, we adapt the linear reconstruction attack approach originally presented by Dwork et al.~\cite{dwork2007price} in this work.

\paragraph{Limitations of prior linear reconstruction approaches.} Linear reconstruction attacks were first presented against data query systems that return noisy answers to queries made over a private dataset~\cite{dinur2003revealing}\footnote{Dinur and Nissim did not explicitly restrict the attention to interactive data query systems. However, their attack as originally proposed assumes that the adversary is able to run queries over known sets of users (the \emph{row-naming} problem). While this assumption is realistic in the context of interactive systems (e.g. those supporting SQL), they do not show how the assumption can be met in context such as pre-defined statistics or synthetic data.}. These attacks require random subsets of users to be selected, as it can typically be done using a unique
user identifier present in data query systems. To our knowledge, the only linear reconstruction attack in a non-interactive setting was proposed by Kasiviswanathan et al.~\cite{kasiviswanathan2013power}, who show that linear reconstruction attacks can theoretically be mounted using $k$-way marginals as well. By recognizing that one of the primary use cases of synthetic data is to preserve $k$-way marginals, we adapt and extend Kasiviswanathan et al.’s attack specifically for synthetic data.

\paragraph{Linear reconstruction attack (\texorpdfstring{$\Advs$}{}).}
\sloppy We refer the reader to Table~\ref{table:notation} which contains a summary of the notation used in the attack. Formally, given the synthetic dataset $\mathcal{S}$, quasi-identifiers of the target record $\mathbf{x}_u$, and quasi-identifiers of all records $X$, the adversary generates a set of $k\text{-way}$ marginal queries  $\mathbf{Q} = \{Q_{\Attr_1, v_1}, \ldots,  Q_{\Attr_q, v_q}\}$, each involving the secret attribute. The adversary then evaluates the noisy answers to these queries from the synthetic dataset $\widehat{\mathbf{r}}$. Then, the adversary solves the following linear program given below which minimizes the total error in all queries ($\sum_{j \in [q]} |\mathbf{e}_j|$). The linear program outputs a vector of reconstructed secret attributes ($\mathbf{t}$) for each record in the target dataset. We set the adversary's guess for the target record's secret attribute as outputs $\widehat{\mathbf{y}'_u} = \lfloor \mathbf{t}_u \rceil$ (nearest integer). \ifusenixfinal In the extended version of the paper~\cite{annamalai2023linear}\else In Appendix~\ref{appsec:varying_k}\fi, we show empirically that $3\text{-way}$ queries provide the best tradeoff between number of queries available to the attack and total noise in the queries.

\vspace{10pt}
\noindent\fbox{%
\parbox{.97\linewidth}{%
\minipage[t]{\dimexpr\linewidth-2\fboxsep-2\fboxrule\relax}
    \textbf{Linear Program.}
    \begin{equation*}
    \begin{split}
        \textbf{variables:}\;\; & \mathbf{t} = (\mathbf{t}_i)_{i \in [n]} \in \mathbb{R}^n \; \text{and}\; \mathbf{e} = (\mathbf{e}_j)_{j \in [q]} \in \mathbb{R}^q \\
        \textbf{minimize:}\;\; & \sum_{j \in [q]} |\mathbf{e}_j| \\
        \textbf{subject to:}\;\; & \forall j \in [q],\; \mathbf{e}_j = \widehat{\mathbf{r}_j} - Q_{\Attr_j, v_j}(\{X | \mathbf{t} \}) \\
        & \forall i \in [n],\; 0 \leq \mathbf{t}_i \leq 1
    \end{split}
    \end{equation*}
\endminipage}
}
\\

Our attack follows the same general structure of Kasiviswanathan et al.'s attack, but we extend it in the following way to improve our attack's success in the SDG setting (see Section~\ref{sec:mar_vs_cond}). Since the adversary is \textit{partially informed}, they can calculate more accurate answers to the 3-way marginal queries by using 3-way \emph{conditional} queries on the synthetic data instead. While marginal queries take the form ``what fraction of people in the dataset are married, employed, and have a college degree?”, conditional queries on the other hand are of the form ``of the people in the dataset that are married and employed, what fraction of them have a college degree?”.

In order to compute the conditional queries, the adversary first generates a set of queries corresponding to the 3-way marginals not involving the secret attribute (``what fraction of people in the dataset are married and employed?”) i.e. $\mathbf{Q}^- = \{Q_{\Attr_1^-, v_1^-}, \ldots,  Q_{\Attr_q^-, v_q^-}\}$ where $|\Attr_i^-| = k - 1$. We then expect $\forall i$, that the fraction:
\[\frac{Q_{\Attr_i, v_i}(\mathcal{D}')}{Q_{\Attr_i^-, v_i^-}(\mathcal{D}')} \approx \frac{Q_{\Attr_i, v_i}(\mathcal{S})}{Q_{\Attr_i^-, v_i^-}(\mathcal{S})}\]
Note that $Q_{\Attr_i^-, v_i^-}(\mathcal{D}') = Q_{\Attr_i^-, v_i^-}(X)$ since the secret attribute is not involved. Furthermore, the adversary can readily calculate this since they have access to the quasi-identifiers of the original dataset, $X$. Therefore $\forall i$:
\[ \widehat{\mathbf{r}}_i = \frac{Q_{\Attr_i, v_i}(\mathcal{S})}{Q_{\Attr_i^-, v_i^-}(\mathcal{S})} \cdot Q_{\Attr_i^-, v_i^-}(X) \]

In Section~\ref{sec:mar_vs_cond}, we show empirically that conditional queries allow the adversary to increase the attack accuracy by up to 10 percentage points compared to marginal queries.

\subsection{Prior attacks on synthetic data}
In this article, we analyze the performance of $\Advs$ and compare it with two prior attacks on synthetic data: distance-based attacks and machine learning inference attacks, respectively introduced by Chen et al.~\cite{chen2020gan} and Stadler et al.~\cite{stadler2022synthetic}.

\paragraph{Distance-based attack ($\Advw$).}
Chen et al.~have proposed a memorization-based heuristic~\cite{chen2020gan}, by measuring the distance between the target record and the closest synthetic record. Intuitively, if the SDG algorithm has in fact memorized parts of the target dataset, then it might tend to output synthetic data samples close to the target dataset.

Formally, given $\mathcal{S}$ and $\mathbf{x}_u$, the adversary outputs
\begin{equation*}
\widehat{\mathbf{y}'_u} = \argmin_{t \in \mathcal{X}_d} \left( \min_{\mathbf{z} \in \mathcal{S}} ||(\mathbf{x}_u | t) - \mathbf{z}||_2 \right)
\end{equation*}

We note that, in practice, for a given partial record $\mathbf{x}_u$, both records $(\mathbf{x}_u | 0),\; (\mathbf{x}_u | 1)$ can be present in the synthetic dataset $\mathcal{S}$. This trivially weakens the adversary, since such a record cannot be attacked (the minimum distance is zero no matter the value $t$). In order to mitigate this scenario, we extend the attack proposed by Chen et al.~\cite{chen2020gan}, using a preprocessing step that collapses records by mode. Before calculating the minimum distances, the adversary groups the records in $\mathcal{S}$ according to the quasi-identifiers and sets the corresponding secret attribute to the most common value (mode) amongst all records with the same quasi-identifiers. If both records appear an equal number of times, the secret attribute is chosen uniformly at random by the adversary.

\paragraph{ML inference attack ($\Advi$).}
Stadler et al.~proposed an attribute inference attack on synthetic data~\cite{stadler2022synthetic} utilizing a classifier $C_\theta : \prod_{i = 1}^{d - 1} \mathcal{X}_i \rightarrow \mathcal{X}_d$ that maps the quasi-identifiers of records to the secret attribute. This classifier is trained on the synthetic data and later applied on the partial target record to retrieve the adversary's guess for the secret bit $\widehat{\mathbf{y}'_u} = C_\theta(\mathbf{x}_u)$.

Intuitively, if the underlying generative model in the SDG algorithm overfits on the training dataset, patterns of association between quasi-identifiers and secret attributes in the training dataset will be reflected in the synthetic data, including for the training record. These patterns of association can then be learned by a classifier to infer the secret attribute of the target record from its quasi-identifiers.

\section{Evaluation Framework}
\label{sec:evaluation}
\subsection{Datasets}
We use two datasets commonly used for synthetic data generation research~\cite{cai2021data, liu2021iterative} and in the NIST competition on synthetic data~\cite{dpsynth2018}: the American Community Survey (ACS) and the San Francisco Fire Department Calls for Service (FIRE).

The ACS dataset, derived from US Census data, is increasingly used in the machine learning community to benchmark new methods~\cite{ding2021retiring} as a modern alternative to the UCI Adult dataset. Specifically, we use the 2018 ACS Employment task for California dataset, used for instance by Liu et al.~to evaluate synthetic data~\cite{liu2021iterative}. This dataset has 16 discrete attributes; we arbitrarily set the secret attribute to be \texttt{SEX}.

The FIRE dataset records fire units' responses to calls~\cite{fire2022}. An earlier version was used for the NIST synthetic data competition~\cite{dpsynth2018,ridgeway2021challenge}. The dataset has a large number of attributes (34). To reduce computational cost, we reduce the dataset to 10 attributes (\texttt{ALS Unit}, \texttt{Call Type Group}, \texttt{Priority}, \texttt{Call Type}, \texttt{Zipcode of Incident}, \texttt{Number of Alarms}, \texttt{Battalion}, \texttt{Call Final Disposition}, \texttt{City} and \texttt{Station Area}) and arbitrarily set the secret attribute to be \texttt{ALS Unit}.

\subsection{Synthetic data generation algorithms}
\label{subsec:sdg_algos}
A large range of computational techniques have been proposed to generate synthetic data from original data, dating back to Rubin’s research on synthetic microdata in 1993~\cite{rubin1993statistical}. Synthetic data generation techniques fall into three categories. Firstly, the traditional statistical paradigm relies on low-dimensional statistical models, including Bayesian networks~\cite{zhang2017privbayes}, Gaussian and vine copulas~\cite{jeong2016copula}, Markov models~\cite{cai2021data}, and decision trees~\cite{reiter2005using}. Secondly, the modern machine learning paradigm relies on high-dimensional deep learning models such as generative adversarial networks (CTGAN) and variational autoencoders (TVAE)~\cite{xu2019modeling}. Finally, the ‘select–measure–generate’ paradigm relies on resampling techniques such as Iterative Proportional Fitting (IPF, also called raking)~\cite{deming1940least}, Relaxed Adaptive Projection (RAP)~\cite{aydore2021differentially}, and Adaptive and Iterative Mechanism (AIM)~\cite{mckenna2022aim}.

\paragraph{Algorithms considered.} To evaluate the success of our attack, we select three widely-used generative models for tabular data, one for each category: BayNet from the traditional statistical paradigm, CTGAN from the machine learning paradigm and RAP from the ‘select–measure–generate’ paradigm.

\paragraph{Baseline algorithms.} We also include two baseline algorithms, NonPrivate (high utility, low privacy) and IndHist\footnote{The IndHist algorithm samples each attribute independently from its respective histogram (see Appendix~\ref{appsec:sdg_algos}). Pairwise associations between attributes are therefore absent from the synthetic dataset, resulting in low utility and high privacy with respect to attribute inference attacks.}~\cite{ping2017datasynthesizer} (low utility, high privacy).

We introduce the NonPrivate algorithm here, while descriptions for the other algorithms included in this study can be found in Appendix~\ref{appsec:sdg_algos}.

The NonPrivate algorithm is a baseline synthetic data generation algorithm, simply sampling from the private dataset with replacement. This can be seen as  the setting where the algorithm has completely overfitted to the private dataset and treats the private dataset as an empirical distribution to sample ``synthetic'' records from. In this setting, the only privacy guarantees for users come from the sampling error inherent to synthetic data generation algorithms.

\subsection{Measuring utility with error metrics}
\label{subsec:err_metrics}
To evaluate the loss in utility between original and synthetic data, we follow standard practices using statistical metrics from the privacy and statistical disclosure control community, used notably by the US Census Bureau, NIST, and EU JRC~\cite{dputil2021,mckenna2021winning,hradec2022multipurpose}. We use two metrics reporting observational error (or measurement error) between marginal distributions of the synthetic data $\mathcal{S}$ and the original data $\mathcal{D}$: the Mean Relative Error (MRE) and the Total Variation Distance (TVD).

Evaluations of utility in privacy-preserving technologies, from de-identification to synthetic data, include a wide range of error metrics~\cite{Wagner2018-me}. Some SDG evaluation studies include metrics reporting the accuracy of machine learning (ML) classification tasks~\cite{tao2021benchmarking}. Due to the complexity of defining a proper classification task (including defining a useful classification target, holdout dataset, hyperparameter search, ML model), we do not include ML metrics in our evaluation. Similar choices were made for the NIST Differentially Private Synthetic Data Competition~\cite{dpsynth2018} and in various works in the area of synthetic tabular data~\cite{mckenna2022aim,mckenna2021winning,aydore2021differentially,liu2021iterative}.

\subsubsection{Mean Relative Error (\texorpdfstring{$\Er$}{})}
To measure the Mean Relative Error (MRE) between marginal queries on original and synthetic data, we follow the practice set forth in the NIST Differentially Private Synthetic Data Challenge~\cite{dpsynth2018} by reporting relative error of random 3-way marginal queries. The MRE is equivalent to the Mean Absolute Percentage Error ($\text{MAPE} = \text{MRE} \times 100$), used notably by the US Census and EU JRC to evaluate the utility of data releases~\cite{uscensus2020,hradec2022multipurpose}.

Since 3-way marginal queries might be often be null or have very small values---which might be suppressed in the synthetic data---we report only the mean relative error on queries that evaluate to more than 10 on the original data ($\text{MRE}_{> 10}$).

\subsubsection{Total Variation Distance (\texorpdfstring{$\Em$}{})}
The Total Variation Distance, also called statistical distance or variational distance, is often used to evaluate the utility of synthetic data~\cite{zhang2017privbayes, cai2021data, mckenna2022aim} and differential private releases of the US Census~\cite{tvdcensus2020}. Informally, it reports the absolute error in histograms calculated from the synthetic data and is bounded between 0 and 1, making it easier to compare errors across different datasets.

\begin{defn}[Total Variation Distance]
Given a fixed subset of attributes $\Attr \subseteq [d]$, $|\Attr| = k$
\begin{equation*}
    \text{TVD}_\Attr(\mathcal{D}, \mathcal{S}) = \frac{1}{2} \sum_{v \in \prod_{i \in \Attr} \mathcal{X}_i} |Q_{\Attr, v}(\mathcal{D}) - Q_{\Attr, v}(\mathcal{S})|
\end{equation*}
\end{defn}

We report the Total Variation Distance averaged across random subsets of $k$ attributes:
\begin{defn}[Average $k$-Total Variation Distance]
Given a set of random subsets of attributes $\Omega = \{\Attr_1, \Attr_2, ..., \Attr_p\}$ s.t. $\forall i\; |\Attr_i| = k$
\begin{equation*}
    k\text{-TVD}(\mathcal{D}, \mathcal{S}) = \frac{1}{p} \sum_{i = 1}^p \text{TVD}_{\Attr_i} (\mathcal{D}, \mathcal{S})
\end{equation*}
\end{defn}
Previous work used both subsets of size $k=3$ and $k=5$~\cite{zhang2017privbayes, cai2021data, mckenna2022aim}; we here select $k=3$ to align with MRE evaluations.

\section{Results}
\label{sec:results}
We evaluate attacks on synthetic data by running the attribute inference privacy game (see Section~\ref{subsubsec:ai_game}) on datasets ACS and FIRE. From each complete dataset, we first sample a dataset $\mathcal{D}$ of size $n=1000$ records. We then run the attack game, generating synthetic datasets using each of the SDG algorithms, and compute the privacy and utility metrics. We average all results (attack accuracy and error metrics) across 500 repetitions of the game. We run our experiments on a single server with a AMD EPYC CPU parallelizing across all 500 games using 32 cores and 500GB of RAM. A detailed analysis of the computational cost of the attack is given in Section~\ref{sec:comp_cost}. The code to reproduce the results is available at \url{https://github.com/synthetic-society/recon-synth}.

\subsection{Evaluation of $\Advs$}
\begin{figure}[htb]
\includegraphics[width=\linewidth]{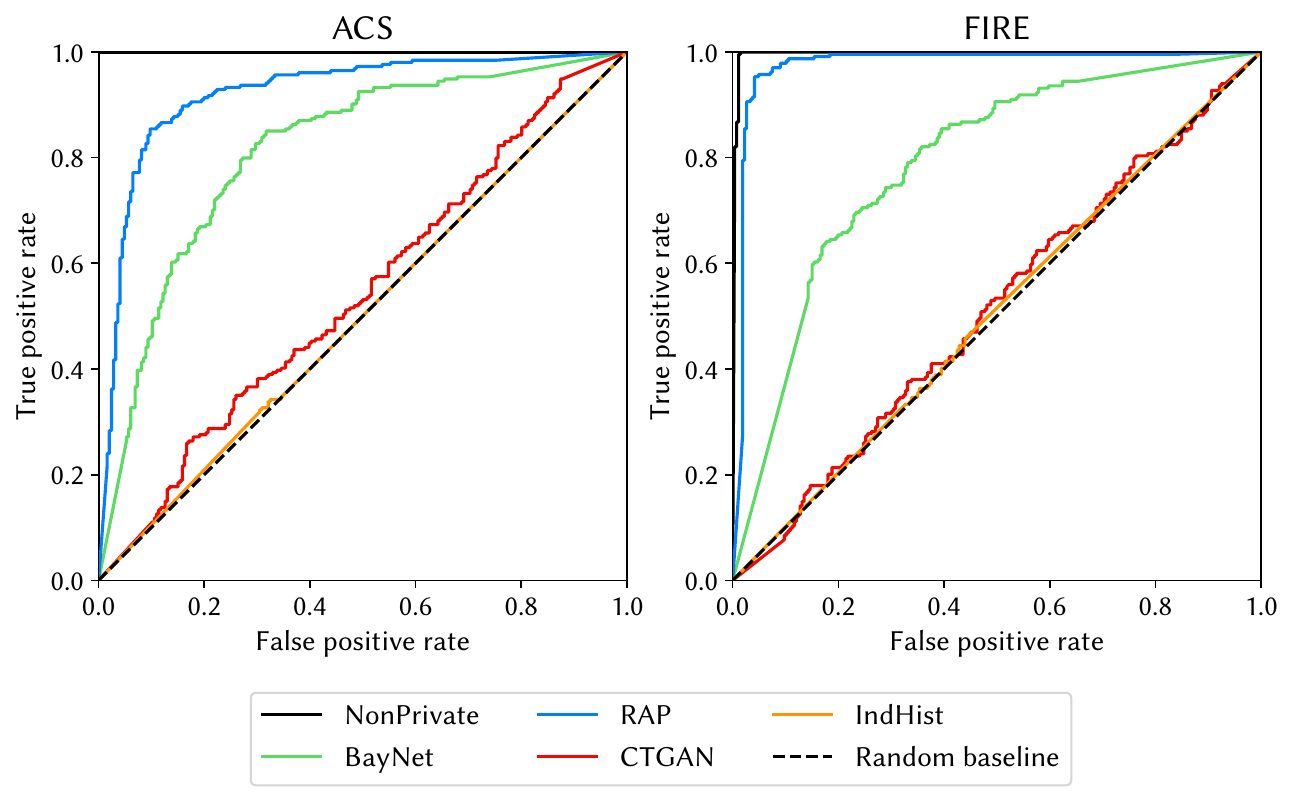}
\caption{Receiver operating characteristic (ROC) curves of $\Advs$ for a synthetic data size $m = 10^6$.}
\centering
\label{fig:privacy_roc}
\end{figure}

We first evaluate the effectiveness of our adversary $\Advs$ when sufficient synthetic data is available. We execute the attribute inference game for a fixed synthetic data size $m = 10^6$, testing $\Advs$ against the six SDG algorithms considered. To compare the estimated score from our attack $\mathbf{t}_u$ with the true secret bit $\mathbf{y}'_u$, we report on Figure~\ref{fig:privacy_roc} the receiver operating characteristic (ROC) curve between $\mathbf{t}_u$ and $\mathbf{y}'_u$.

Our attack achieves a consistently high AUC ($> 0.75$) for RAP and BayNet algorithms, as well as the NonPrivate baseline, across both datasets. The attack is not effective on CTGAN and IndHist, achieving an AUC close to a random prediction ($\text{AUC} \approx 0.5$). This is expected for IndHist, which should be protected against most attribute inferences since any association between attributes is lost. While the low AUC on CTGAN and IndHist suggests resistance to such attribute inference attacks, this comes at a significant cost to utility as will be shown in Section~\ref{subsec:synth_data_size}.

\subsection{Comparison to prior attacks}
\begin{figure}[htb]
\includegraphics[width=\linewidth]{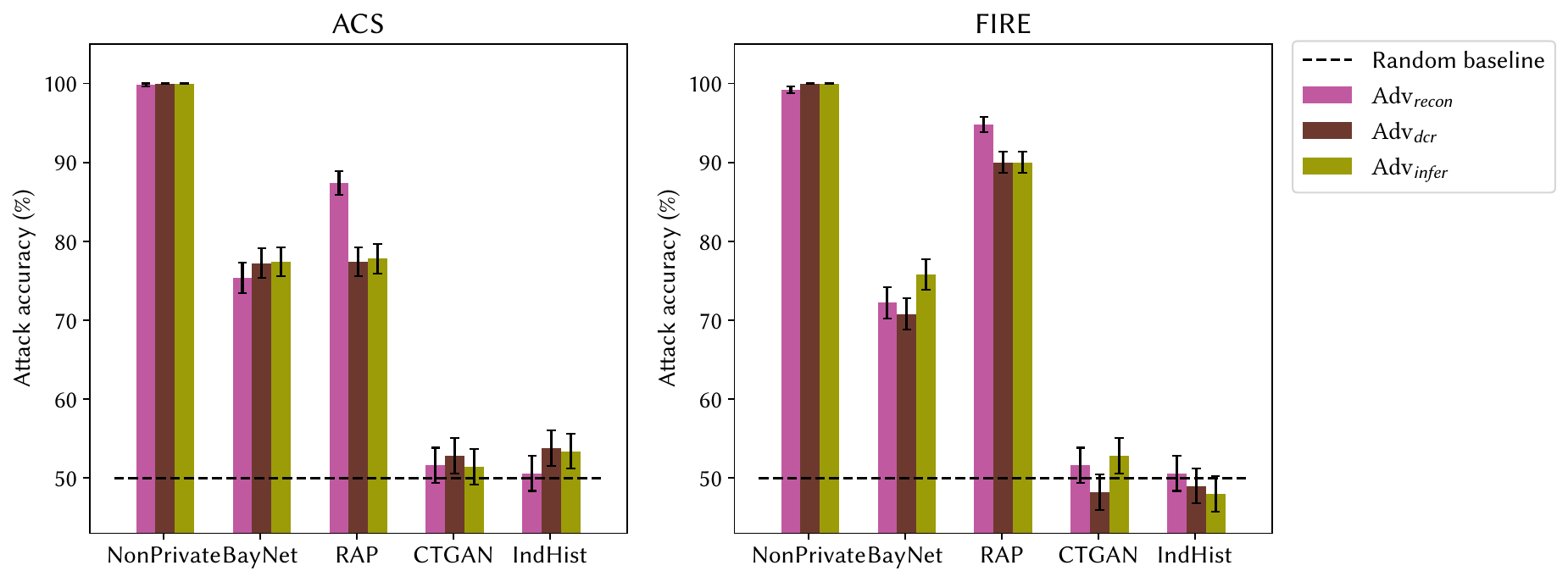}
\caption{Comparison of attack accuracy (mean $\pm$ s.d.) between $\Advs$ and prior attacks $\Advw$ and $\Advi$ for a synthetic data size $m = 10^6$.}
\centering
\label{fig:recon_vs_dcr}
\end{figure}

We then compare our attack with two prior attacks ($\Advw$ and $\Advi$) from the literature. 
Figure~\ref{fig:recon_vs_dcr} reports the accuracy between $\mathbf{y}'_u$ and $\widehat{\mathbf{y}'}_u$ for each attack, for a synthetic data size $m=10^6$ (see \ifusenixfinal extended version of the paper~\cite{annamalai2023linear} \else Appendix~\ref{appsec:recon_vs_dcr_full} \fi for $m = 10$ to $10^6$).

Our adversary outperforms or performs statistically close (within 2 standard deviations) to $\Advw$ and $\Advi$ for all SDG algorithms. In particular, our attack achieves a much higher accuracy on RAP than previous attacks, with a 9.60 and 4.80 percentage point (p.p.) increase in attack accuracy over the memorization-based and inference attacks for the ACS and FIRE datasets respectively. This suggests that state-of-the-art attacks can potentially underestimate the privacy risks of synthetic data specifically in the `select-measure-generate' paradigm. Lastly, for the CTGAN and IndHist algorithms, we notice that even when a synthetic dataset with a large number of records ($m = 10^6$) is released, none of the adversaries are able to perform significantly better than the random baseline.

\subsection{Impact of synthetic data size}
\label{subsec:synth_data_size}

We have so far evaluated attack accuracy when sufficient synthetic data is released ($m=10^6$). However, data releases can in practice be restricted to small samples. To evaluate if attacks still succeed when fewer synthetic records are released, we test our attack on synthetic datasets from $m=10$ to $m=10^6$.

\begin{figure}[bht]
\includegraphics[width=\linewidth]{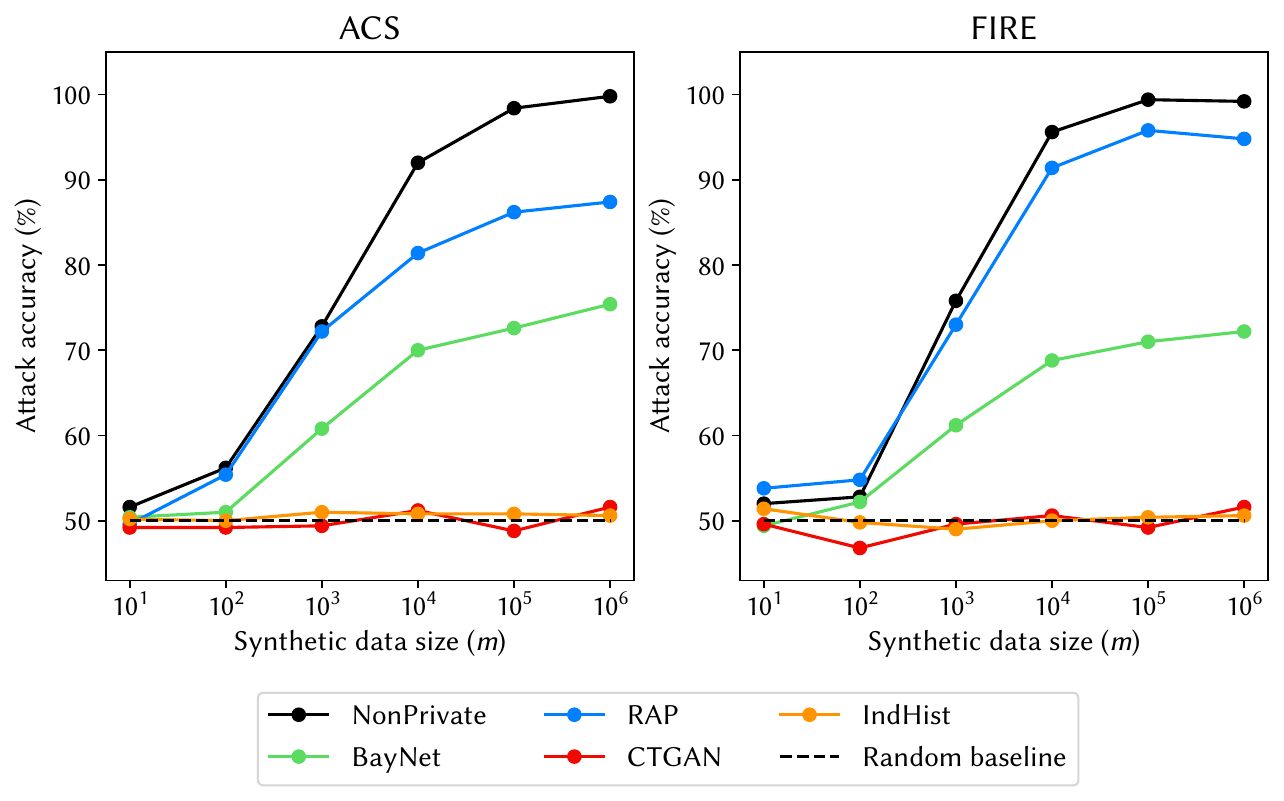}
\caption{Attack accuracy of $\Advs$ for a synthetic data size from $m=10$ to $10^6$.}
\centering
\label{fig:privacy}
\end{figure}

Figure~\ref{fig:privacy} shows that the accuracy of our attack is low for $m=10^2$ and increases with $m$ against NonPrivate, BayNet and RAP. The accuracy is close to 50\% for CTGAN and IndHist for all values of $m$. The attack accuracy is 87.4\% and 94.8\% against RAP on resp. ACS and FIRE datasets, at $m=10^6$. In contrast, this accuracy is reduced by resp. 15.2 and 21.8 p.p. when evaluating the attack on a small synthetic dataset of size $m=10^3$. When even fewer records are released, the attack accuracy is further reduced, resulting in low accuracy of resp. 55.4\% and 54.8\% against RAP at $m = 10^2$. We observe that similar trends exist for the BayNet algorithm as well: a decrease of 14.6 and 11.0 p.p. in accuracy from $m = 10^6$ to $m = 10^3$, for the ACS and FIRE datasets respectively.

This suggests that benchmark evaluations on small synthetic datasets can understate the practical risks of attribute inference. As a result, we argue that while it may be common currently to only evaluate the privacy risks for the specific synthetic data size that is to be released, privacy evaluations of SDGs should potentially include evaluations on larger synthetic data sizes as well in order to achieve a more comprehensive and robust understanding of the privacy leakage from SDGs. We note, however, that the association between synthetic data size and attack accuracy is not trivial: too large synthetic data sizes could, for instance, result in overflow errors or reach a limit in computational power that may, in turn, lower attack accuracy.

\begin{figure}[htb]
\includegraphics[width=0.85\linewidth]{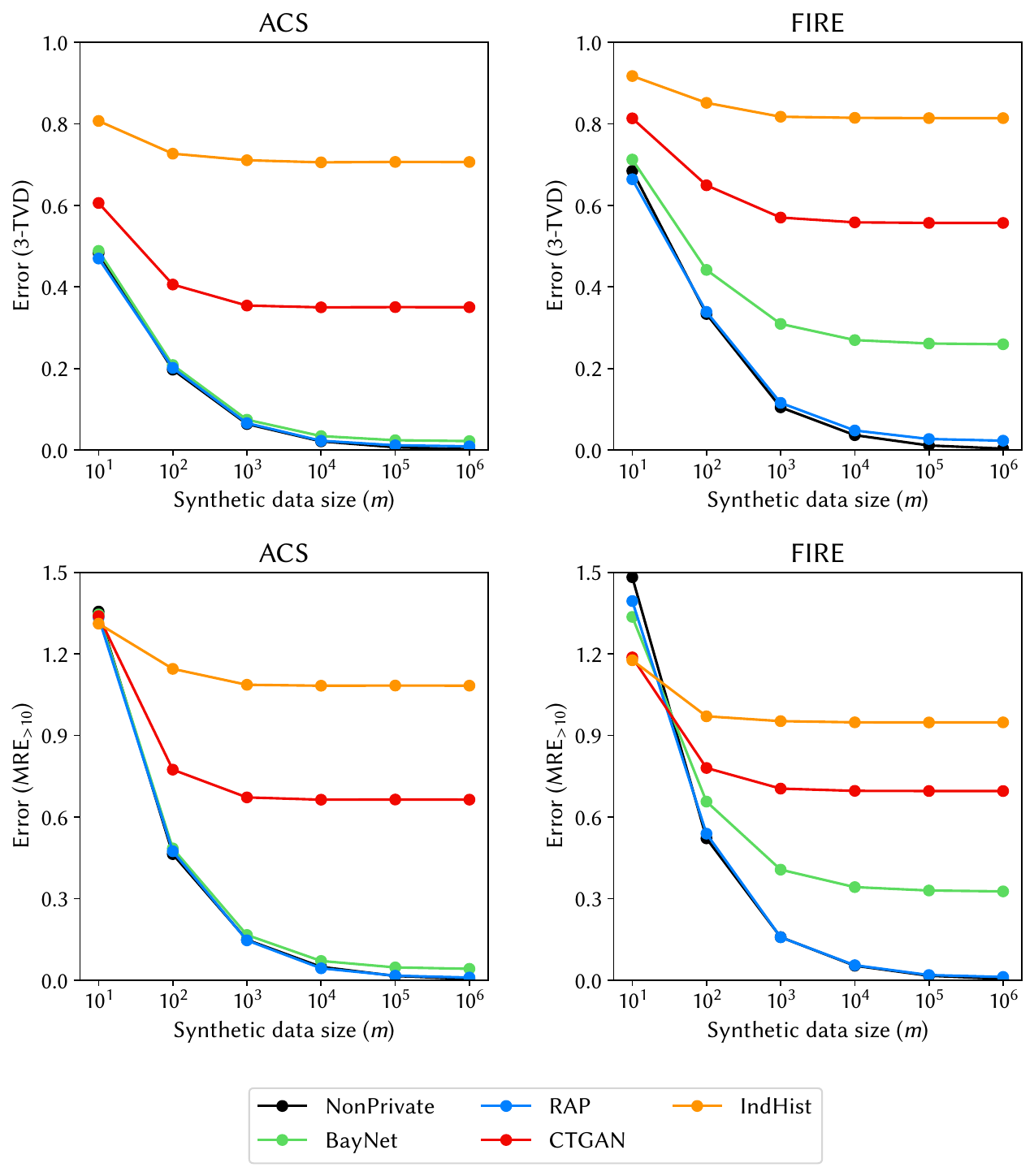}
\centering
\caption{Measurement errors for synthetic data sizes $m=10$ to $10^6$.}
\label{fig:utility}
\end{figure}

While releasing smaller synthetic data can limit the likelihood of inference attack, this comes at a high cost in terms of utility. Figure~\ref{fig:utility} reports the utility error decreases fast as the synthetic data size increases (see \ifusenixfinal extended version of the paper~\cite{annamalai2023linear} \else Appendix~\ref{appsec:exact_error} \fi for exact error values). This is a natural consequence of the sampling error (the more records, the more accurate statistics---as demonstrated by NonPrivate). The error metric appears to asymptotically reach zero for NonPrivate and RAP only. This suggests that RAP is, in our experiment, the only SDG algorithm that provides low bias. The error metrics on IndHist, BayNet, and CTGAN plateau often or always, suggesting evidence of model bias. This is expected for the IndHist baseline: after $m=10^3$, all univariate marginals are correctly estimated, yet the error remains high due to the absence of higher-order associations.

\subsection{Privacy-utility tradeoff}

Our findings show that protection against inference attacks and utility of synthetic data are both conditioned on multiple factors such as the SDG algorithm, the synthetic data size, and the type of attack. To understand if SDG algorithms can provide sufficient utility while protecting against attribute inference, we systematically evaluate the tradeoffs between attack accuracy and measurement error.

\begin{figure}[htb]
\includegraphics[width=\linewidth]{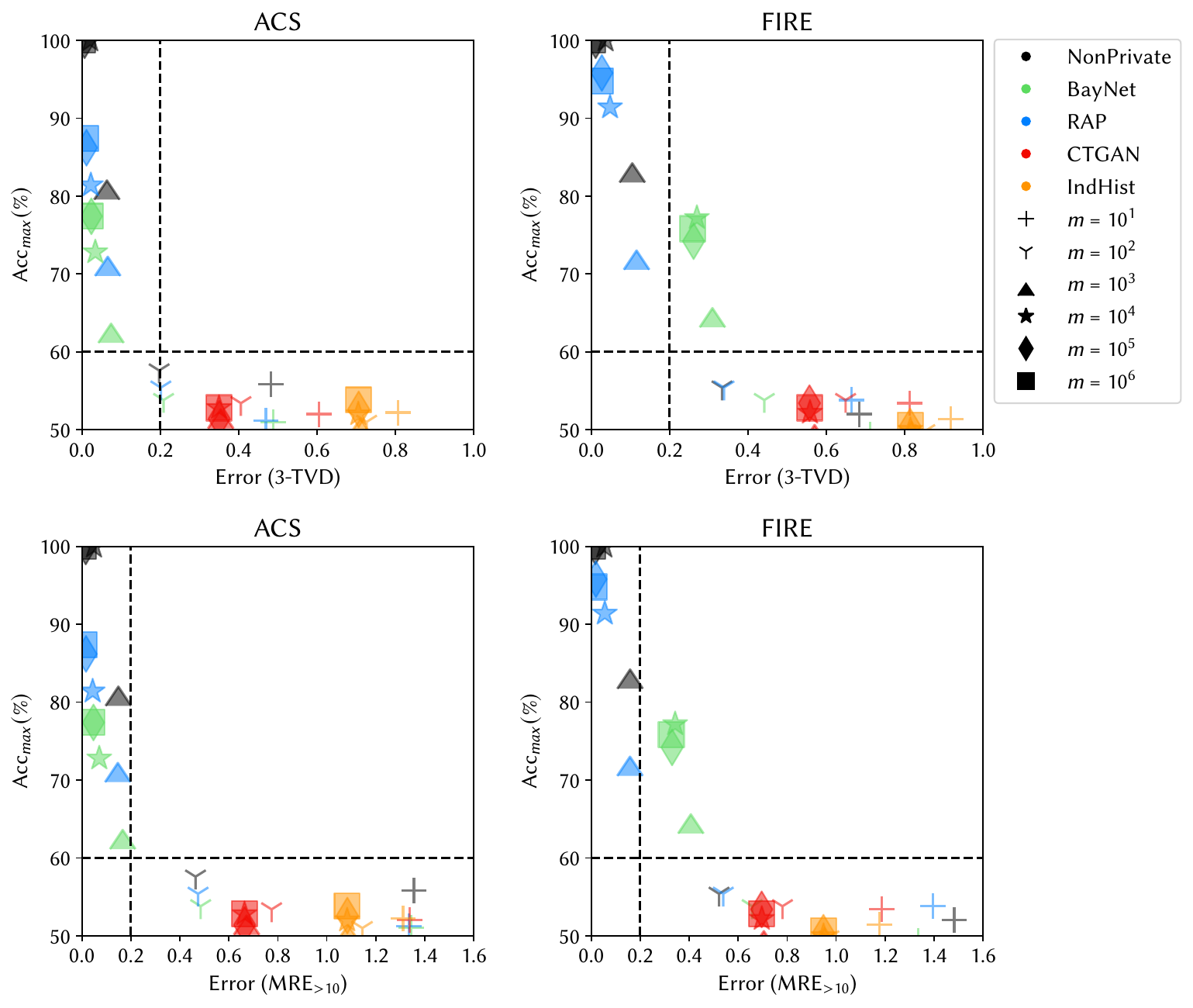}
\caption{Tradeoff plot between privacy (attack accuracy) and utility (measurement error).}
\centering
\label{fig:priv_utility_tradeoff}
\end{figure}

Figure~\ref{fig:priv_utility_tradeoff} shows the tradeoff between privacy and utility achieved by all models for synthetic data sizes from $m=10$ to $10^6$. For each data point, we report the maximum attack accuracy across the three attacks ($\Advs$, $\Advi$, and $\Advw$):

\begin{equation*}
    \Amax = \max_{\Adv \in \{\Advs, \Advw, \Advi\}} \text{Accuracy}^\Adv
\end{equation*}

The maximum accuracy represents an upper-bound empirical estimate, and may not be reached by an adversary in practice. However, it can be a useful worst-case metric when evaluating privacy-utility tradeoffs empirically (tradeoffs for each attack separately are reported in \ifusenixfinal the extended version of the paper~\cite{annamalai2023linear}\else Appendix~\ref{appsec:priv_util_tradeoffs})\fi).

To help visualise these tradeoffs, we set baseline rates at $\Amax > 60\%$ (high attack accuracy, y-axis) as well as $\Er < 0.20$ and $\Em < 0.20$ (low error rate, x-axis). We stress, however, that there is no consensus on which attack accuracy above 50\% in a game would be sufficient to significantly breach privacy in practice. Even at $\Amax = 50\%$, some outlier records may still be at risk~\cite{stadler2022synthetic}. Similarly, an average relative error rate of 0.20 might, for instance, be too high for statistical analyses in low-powered study and result in incorrect estimates of effect, especially at small synthetic data sizes~\cite{loken2017measurement}.

Firstly, we observe that whenever the synthetic data has high utility (low error), it also offers little to no privacy protection. On the ACS dataset, when $\Er < 0.20 $ and $\Em < 0.20$, the lowest $\Amax$ is 60.8\% for BayNet (at $m=10^3$), and 65.4\% for RAP (at $m=10^3$). Similarly, on the FIRE dataset, when $\Er < 0.20 $ and $\Em < 0.20$, the lowest $\Amax$ is 68.6\% for RAP (at $m=10^3$). On both datasets, at all synthetic data sizes $m$ considered, CTGAN never yields an error metric $\Er$ or $\Em$ below $0.20$.

On the other hand, whenever synthetic data offers strong privacy protection, this results in low utility with large errors. On the ACS dataset, when $\Amax < 60\%$, the lowest error $\Er$ is 0.664 for CTGAN (at $m=10^4$), 0.485 for BayNet (at $m=10^2$), and 0.474 for RAP (at $m=10^2$). Similarly, on the FIRE dataset, when $\Amax < 60\%$, the lowest error $\Er$ is 0.696 for CTGAN (at $m=10^6$), 0.657 for BayNet (at $m=10^2$), and 0.539 for RAP (at $m=10^2$). These high error rates would prevent most research tasks on these two datasets.

In all our experiments, we could not find a setting where any of the considered SDG algorithms both provide sufficient privacy protection ($\Amax < 60\%$) as well as utility that would enable some research tasks ($\Er < 0.20$ and $\Em < 0.20$). 

\subsection{Impact of differential privacy}
Differential privacy~\cite{dwork2006calibrating} is a popular approach to preventing privacy leakages that provides strong theoretical privacy guarantees by design.  Differentially private mechanisms are defined using a parameter $\varepsilon$ which controls the privacy-utility tradeoff. This parameter, also referred to as the privacy budget, ranges from 0 (perfect protection) to $\infty$ (no added protection).

Intuitively, differential privacy guarantees that adding, removing, or changing the content of an individual record will not affect much the (distribution of) synthetic datasets that are generated. In our attribute inference privacy game, the challenger randomizes the target record's secret attribute (see Section~\ref{sec:discussion}). Differential privacy should therefore significantly reduce the attack accuracy of any potential adversary. At the same time, research shows that differentially private mechanisms can also strongly limit utility and affect model performance in disparate and unpredictable ways~\cite{tramer2021differentially,ganev2022robin,bagdasaryan2019differential,stadler2022synthetic}. Below, we investigate the privacy-utility tradeoffs of differential privacy in the context of synthetic data and attribute inference.

To understand if implementing DP can help provide a better privacy-utility tradeoff, we replicate our experiments of the SDG algorithms with their DP equivalents. Out of the SDG algorithms considered, three have differentially private equivalents---$\RAPDP$~\cite{liu2021iterative} for RAP, PrivBayes~\cite{zhang2017privbayes} for BayNet, and DPCTGAN~\cite{rosenblatt2020differentially} for CTGAN (see to Appendix~\ref{appsec:sdg_algos} for details about each algorithm). Since CTGAN does not provide sufficient utility in our experiments, we restrict ourselves to evaluating only $\RAPDP$ and PrivBayes.

We note that $\RAPDP$ and PrivBayes satisfy two different variants of differential privacy. $\RAPDP$ satisfies a relaxation known as approximate or $(\varepsilon, \delta)$-DP~\cite{liu2021iterative}. PrivBayes satisfies the original $\varepsilon$-DP~\cite{zhang2017privbayes}. In our evaluation, we set $\delta = \frac{1}{n^2} = 10^{-6}$ for $\RAPDP$, using the same value set by Liu et al.~\cite{liu2021iterative}.

\begin{figure}[htb]
\includegraphics[width=\linewidth]{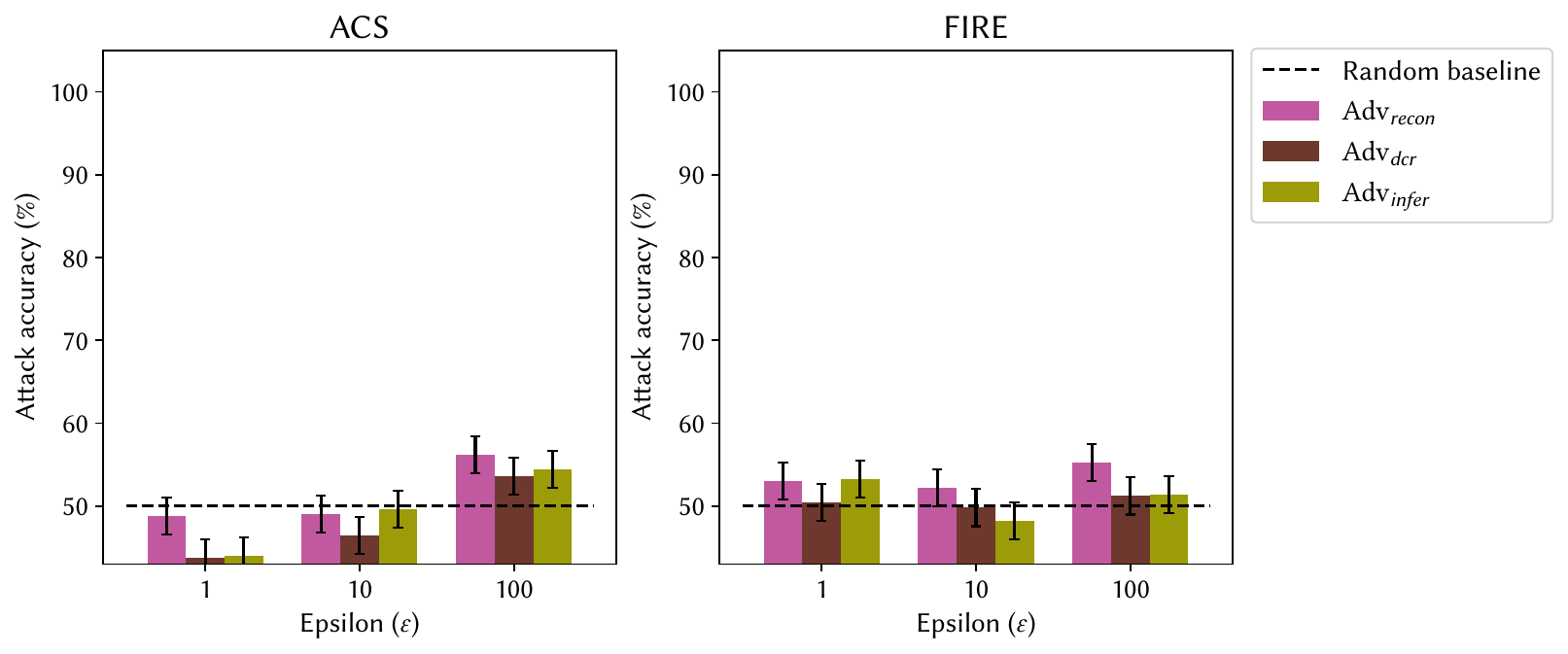}
\caption{Attack accuracy of $\Advs$, $\Advw$ and $\Advi$ as a function of privacy parameter $\varepsilon$ for $\RAPDP$ synthetic data of size $m = 10^6$}
\centering
\label{fig:privacy_dp_rapdp}
\end{figure}

\begin{figure}[htb]
\includegraphics[width=\linewidth]{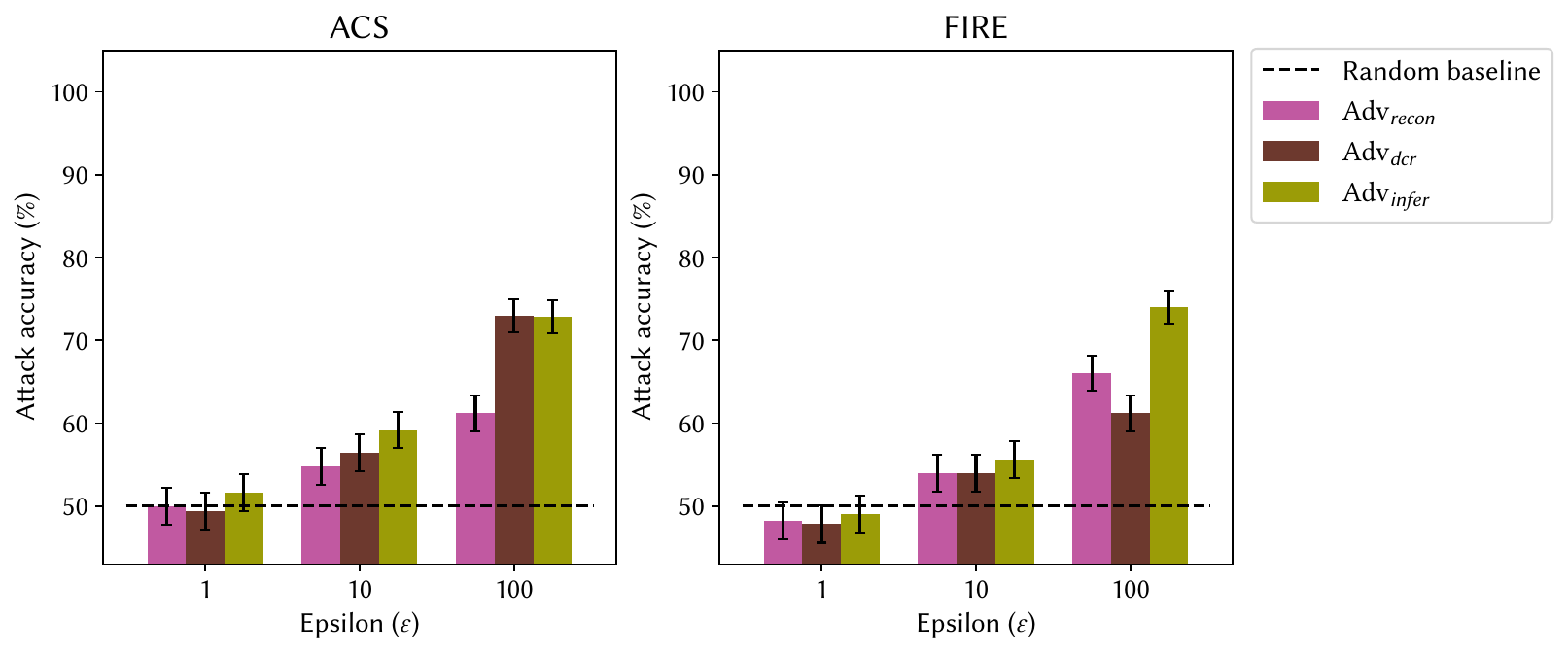}
\caption{Attack accuracy of $\Advs$, $\Advw$ and $\Advi$ as a function of privacy parameter $\varepsilon$ for PrivBayes synthetic data of size $m = 10^6$}
\centering
\label{fig:privacy_dp_privbayes}
\end{figure}

\paragraph{Impact of privacy budget.} We first evaluate the impact of the privacy budget $\varepsilon$ on attack accuracy. Figure~\ref{fig:privacy_dp_rapdp} and Figure~\ref{fig:privacy_dp_privbayes} show the accuracy of all three attacks ($\Advs$, $\Advi$ and $\Advw$) increases with $\varepsilon$ for both $\RAPDP$ and PrivBayes. We observe that differential privacy has a large impact on attack accuracy, even for a moderately large privacy budget $\varepsilon = 10$. DP mechanisms severely reduce the attack accuracy for all three adversaries compared to their non-differentially private counterparts, across both datasets. For example, for the RAP algorithm on the ACS dataset, the accuracy of $\Advs$, $\Advw$ and $\Advi$ are reduced by resp. 38.4, 31.0, and 28.2 percentage points (p.p.) by using $\RAPDP$ at $\varepsilon = 10$. This suggests that mechanisms with moderately large privacy parameters could significantly reduce the risk of attribute inference attacks against realistic adversaries.

\begin{figure}[H]
\includegraphics[width=0.85\linewidth]{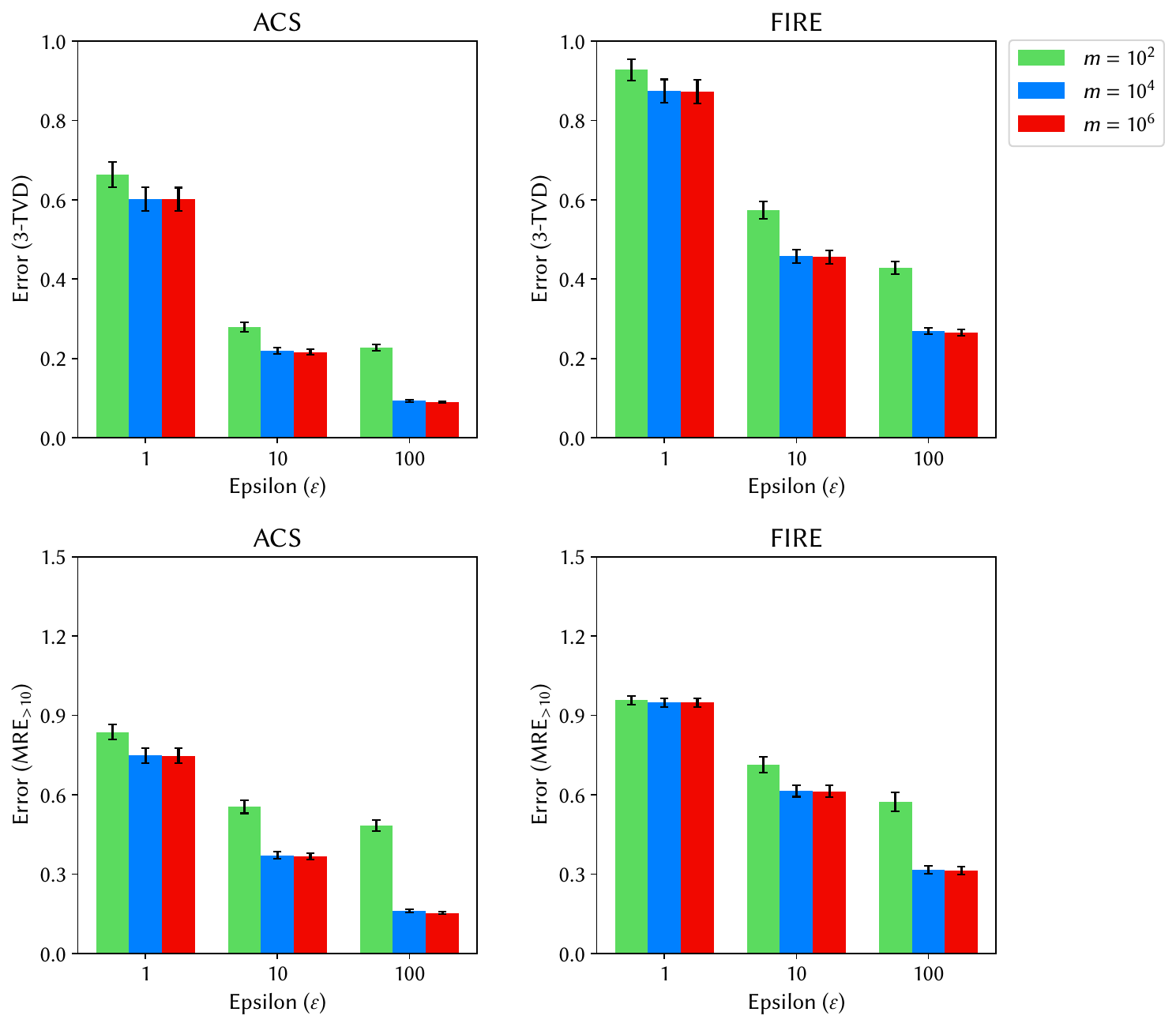}
\centering
\caption{Error as a function of privacy parameter $\varepsilon$ for $\RAPDP$}
\label{fig:utility_dp_rapdp}
\end{figure}

\begin{figure}[H]
\centering
\includegraphics[width=0.85\linewidth]{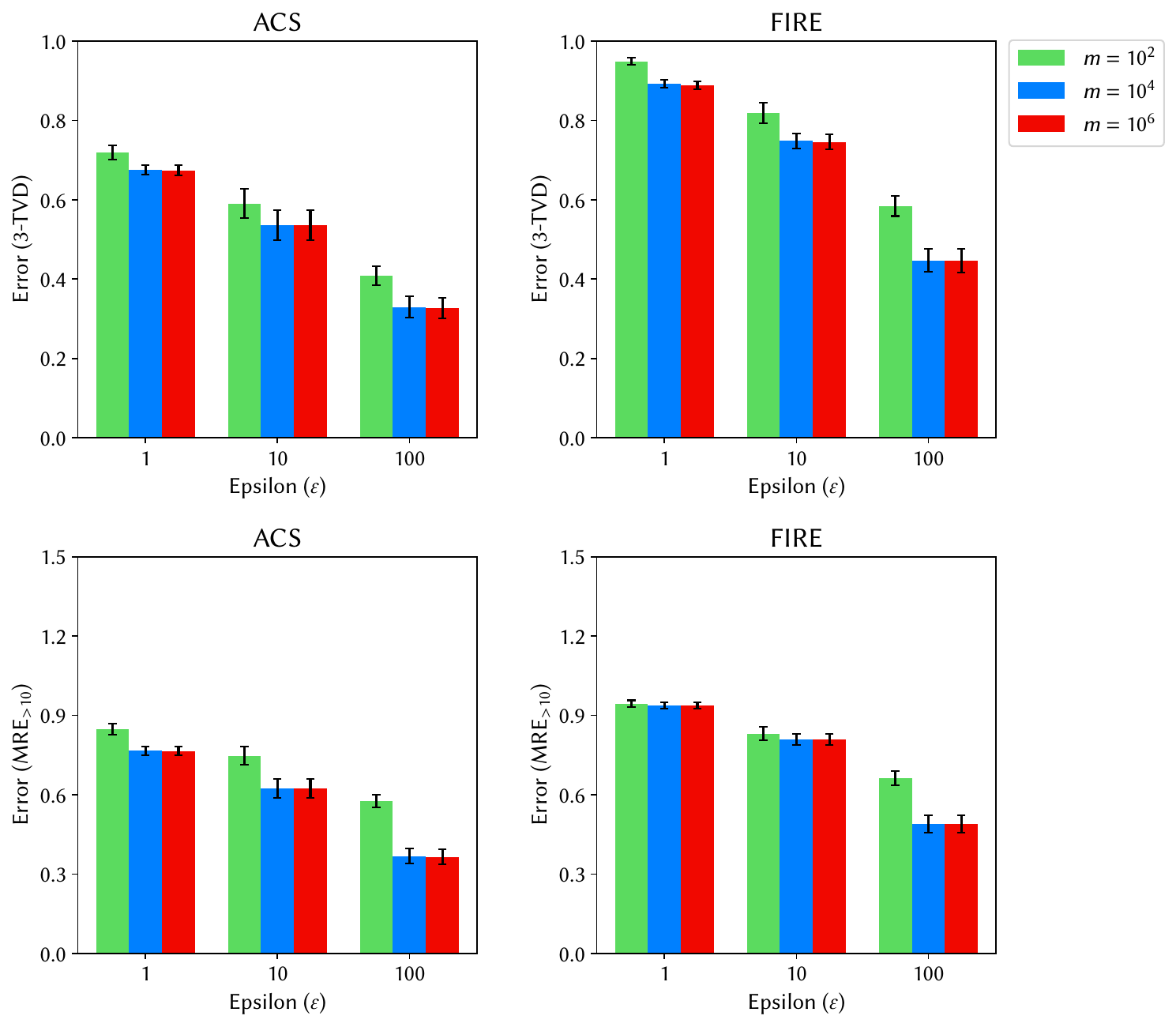}
\caption{Error as a function of privacy parameter $\varepsilon$ for PrivBayes}
\label{fig:utility_dp_privbayes}
\end{figure}

At the same time, DP mechanisms are known to significantly impact utility~\cite{kenny2021use,cummings2023challenges}. We therefore evaluate the effect of the privacy budget $\varepsilon$ on the two error metrics in Figure~\ref{fig:utility_dp_rapdp} and Figure~\ref{fig:utility_dp_privbayes} (see \ifusenixfinal extended version of the paper~\cite{annamalai2023linear} \else Appendix~\ref{appsec:exact_error_dp_rapdp} and Appendix~\ref{appsec:exact_error_dp_privbayes} \fi for exact values) for the $\RAPDP$ and PrivBayes algorithms respectively. We find that at each $\varepsilon$ level, $\RAPDP$ provides better utility (lower error) than PrivBayes, on both metrics and for both datasets ACS and FIRE.

We observe that, unlike the non-DP synthetic data generation algorithms, the utility of $\RAPDP$ and PrivBayes is not as strongly dependent on the size of the synthetic data released. For example, the relative error on the RAP and BayNet algorithms decreases by resp. 0.46 and 0.44, when releasing ACS synthetic data of size $10^6$ compared to $10^2$. However, the drop in relative error for the $\RAPDP$ and PrivBayes algorithms is at most 0.33 and 0.21 resp. (achieved at $\varepsilon = 100$).

\begin{figure}[H]
\includegraphics[width=\linewidth]{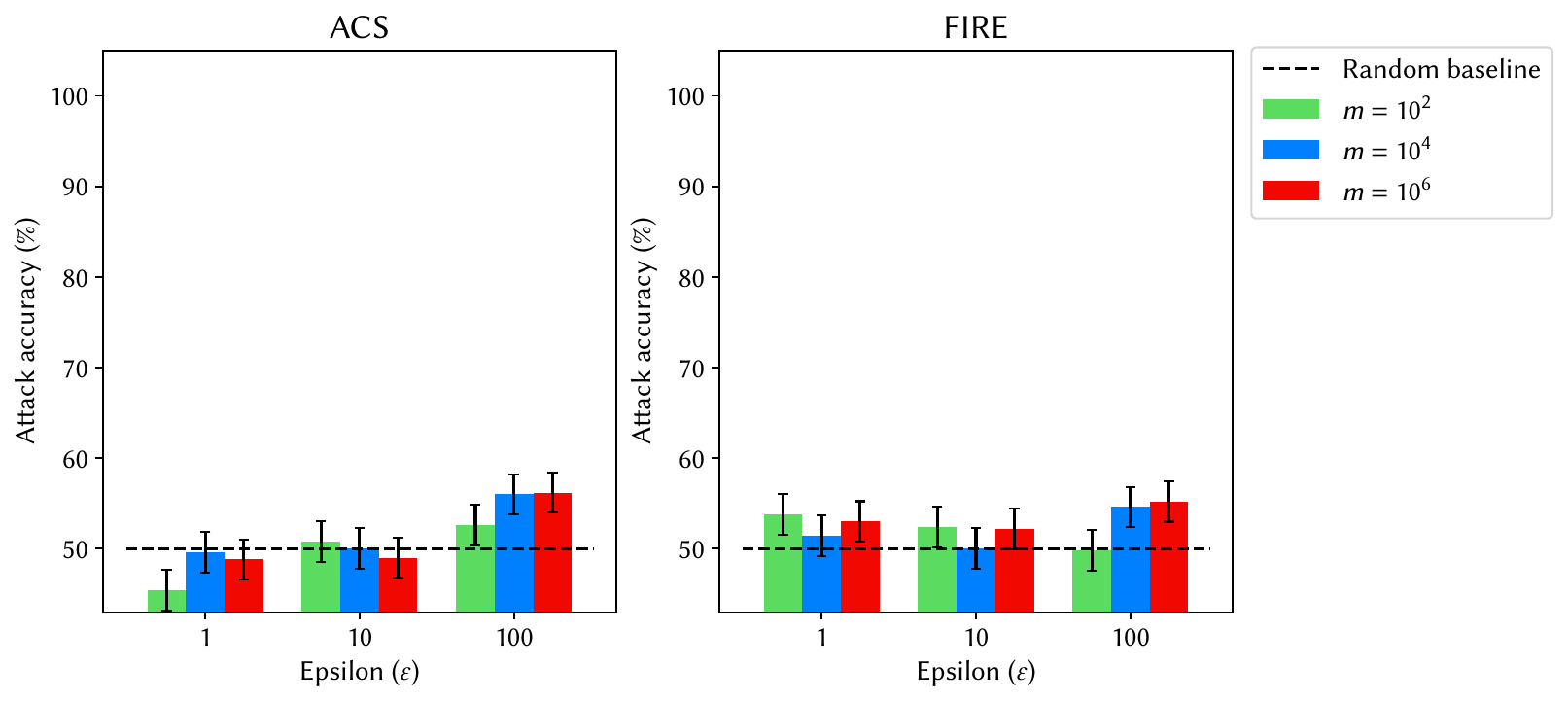}
\caption{Attack accuracy of $\Advs$ as a function of privacy parameter $\varepsilon$ for $\RAPDP$}
\centering
\label{fig:privacy_synth_size_dp_rapdp}
\end{figure}

\begin{figure}[H]
\includegraphics[width=\linewidth]{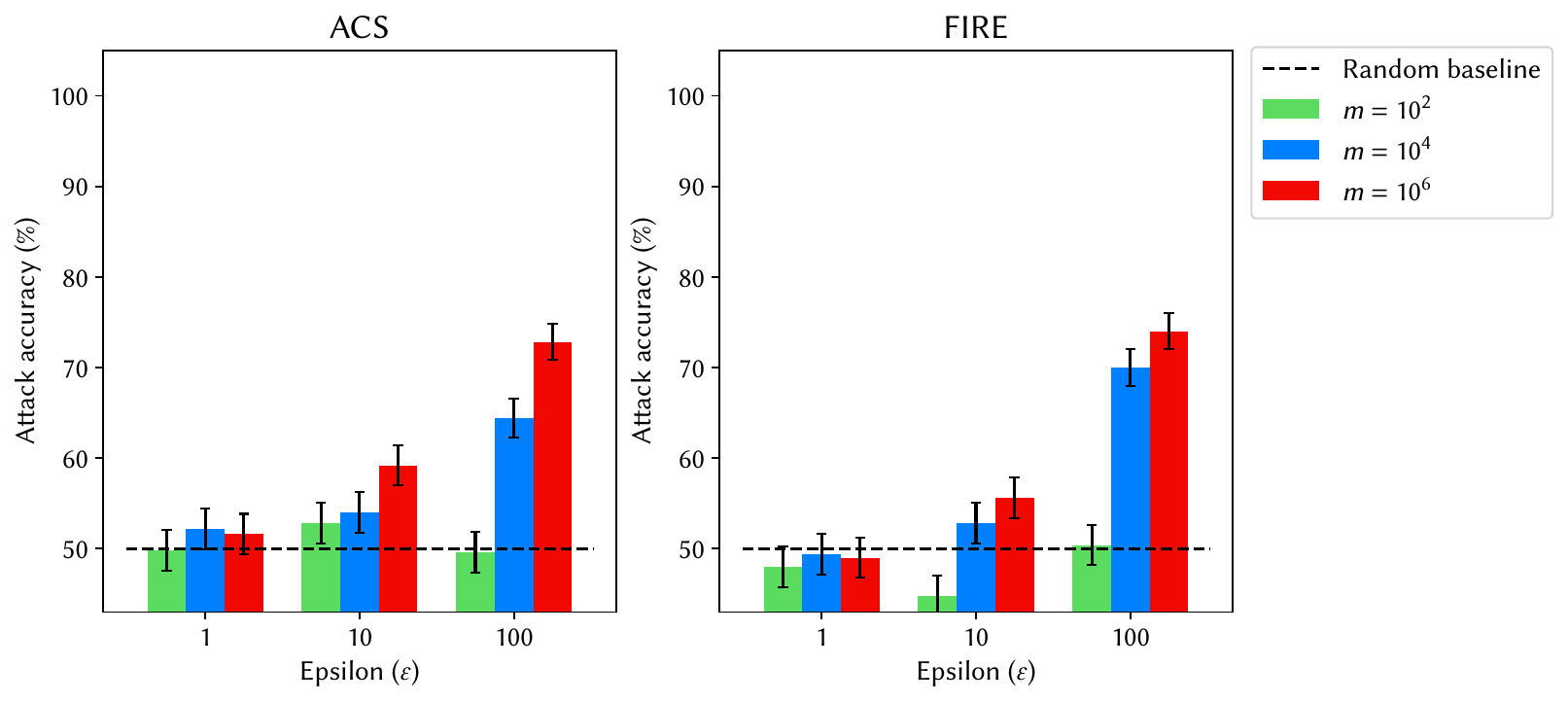}
\caption{Attack accuracy of $\Advi$ as a function of privacy parameter $\varepsilon$ for PrivBayes}
\centering
\label{fig:privacy_synth_size_dp_privbayes}
\end{figure}

\paragraph{Impact of synthetic data size.} $\Advs$ achieves higher accuracy than all other adversaries for $\RAPDP$. Similarly, $\Advi$ achieves higher accuracy for PrivBayes. We therefore show the accuracy of $\Advs$ and $\Advi$ respectively for three synthetic data sizes ($m = 10^2$, $10^4$ and $10^6$) in Figure~\ref{fig:privacy_synth_size_dp_rapdp} and Figure~\ref{fig:privacy_synth_size_dp_privbayes}. We note that for moderate to large privacy parameters ($\varepsilon \geq 10$), the synthetic data size heavily impacts the accuracy of $\Advi$ whereas the accuracy of $\Advs$ is no longer impacted by the synthetic data size even at $\varepsilon = 100$. Similar to our earlier finding, we see here that, at moderate to large privacy parameters, there can be up to a 23.2 p.p. drop in attack accuracy when sampling a small synthetic dataset of size $10^2$ (achieved for $\Advi$ on PrivBayes at $\varepsilon = 100$). This confirms that, for certain SDG algorithms, the synthetic data size is ultimately an important factor when evaluating practical privacy risks of synthetic data, even when differentially private noise is added.

\begin{figure}[thb]
\includegraphics[width=\linewidth]{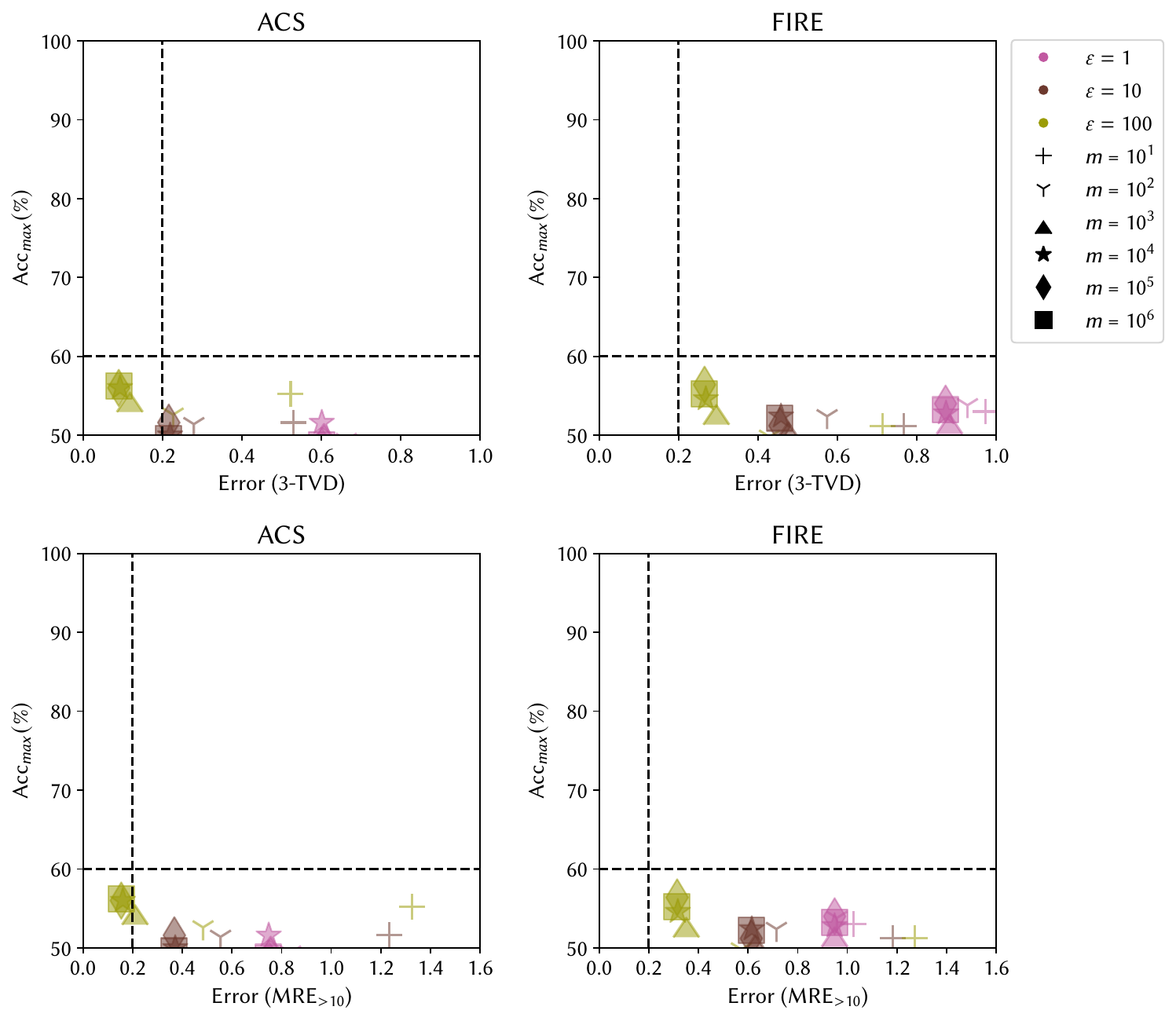}
\caption{Tradeoff between privacy and utility for $\RAPDP$ at $\varepsilon$ = 1, 10 and 100}
\centering
\label{fig:priv_utility_tradeoff_dp_rapdp}
\vspace{-0.5cm}
\end{figure}

\begin{figure}[thb]
\includegraphics[width=\linewidth]{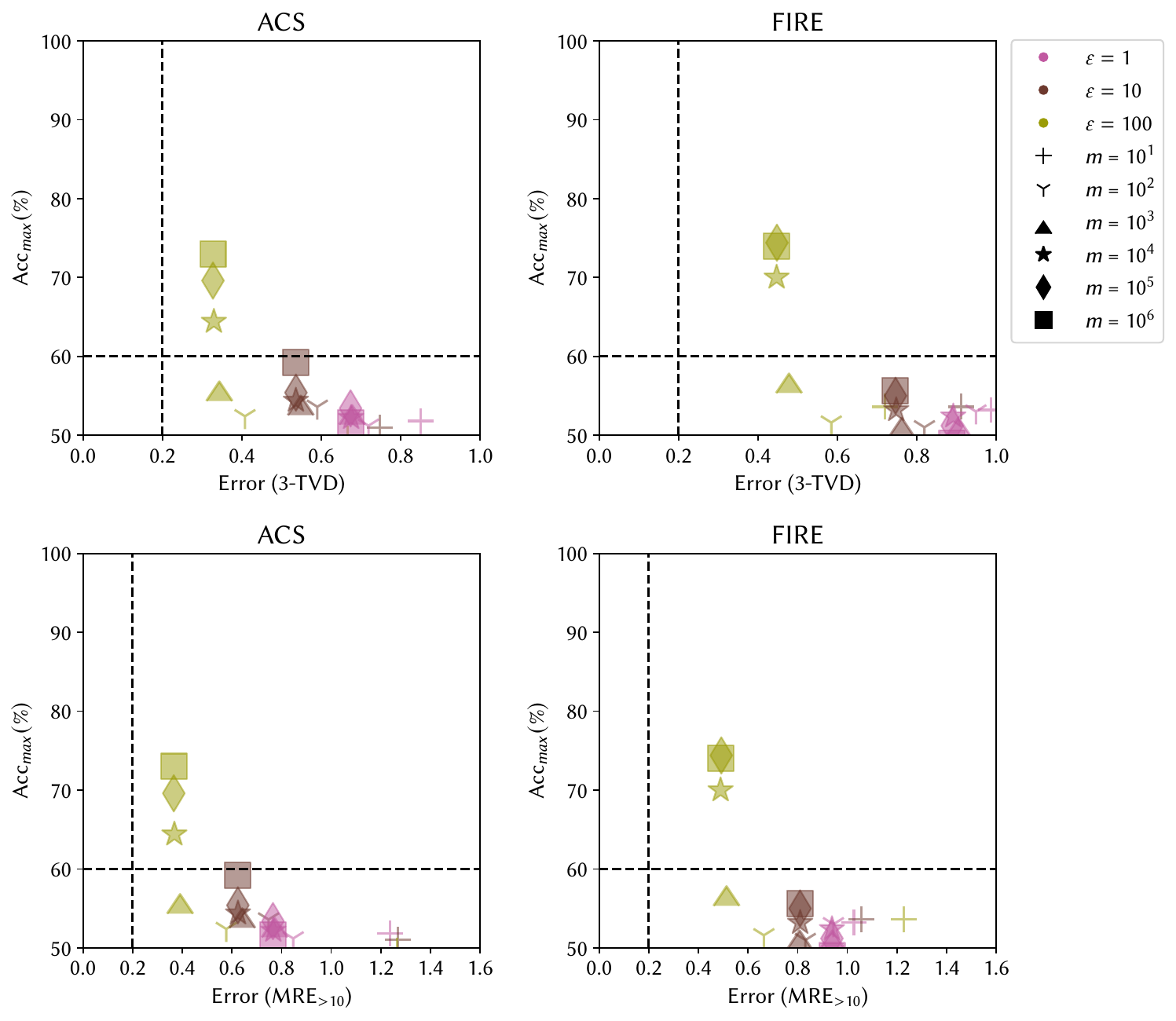}
\caption{Tradeoff between privacy and utility for PrivBayes at various $\varepsilon$ = 1, 10 and 100}
\centering
\label{fig:priv_utility_tradeoff_dp_privbayes}
\end{figure}

\paragraph{Privacy-utility tradeoffs.} Figure~\ref{fig:priv_utility_tradeoff_dp_rapdp} and Figure~\ref{fig:priv_utility_tradeoff_dp_privbayes} show the attack accuracy and measurement errors of the two DP SDG algorithms, across datasets and synthetic data sizes. For PrivBayes, we note that the privacy gains achieved by differential privacy guaranteed are subject to similar drops in utility as for BayNet. Across all datasets and error metrics, the PrivBayes algorithm is not significantly closer to the bottom left-hand corner than the BayNet algorithm.

On the other hand, $\RAPDP$ at $\varepsilon = 100$ can in fact produce synthetic data that has better combination of utility and privacy compared with non-DP methods we previously evaluated (see Fig~\ref{fig:priv_utility_tradeoff}), but only on the ACS dataset. Synthetic data generated by $\RAPDP$ at $\varepsilon = 100$ is closer to the bottom left-hand corner than that generated by the RAP algorithm. This suggests that, although moderate to large values of $\varepsilon$ may not provide strong theoretical protections, the way these guarantees are implemented can help provide slightly better privacy-utility tradeoffs against practical attacks.

\subsection{Impact of choice and number of queries}
\label{sec:mar_vs_cond}
In previous sections, we have analyzed the impact of \textit{external} factors such as the SDG algorithm, synthetic data size, and differential privacy defenses on the success of our attack. Here, we further analyze the impact of the \textit{internal} parameters used to define the attack and provide an explanation for why our attack works. Specifically, we focus on the two main parameters that enable our attack to achieve good accuracy---the choice and number of queries. In this section, we consider synthetic data of size $m = 10^6$ and the FIRE dataset (see \ifusenixfinal extended version of the paper~\cite{annamalai2023linear} \else Appendix~\ref{appsec:varying_k} and~\ref{appsec:varying_n_queries} \fi for a more in-depth analysis). We do not include CTGAN and IndHist as these factors do not affect the attack accuracy since even in the best case, the attack accuracy for CTGAN and IndHist is close to the random baseline.

\begin{figure}[htb]
\includegraphics[width=0.9\linewidth]{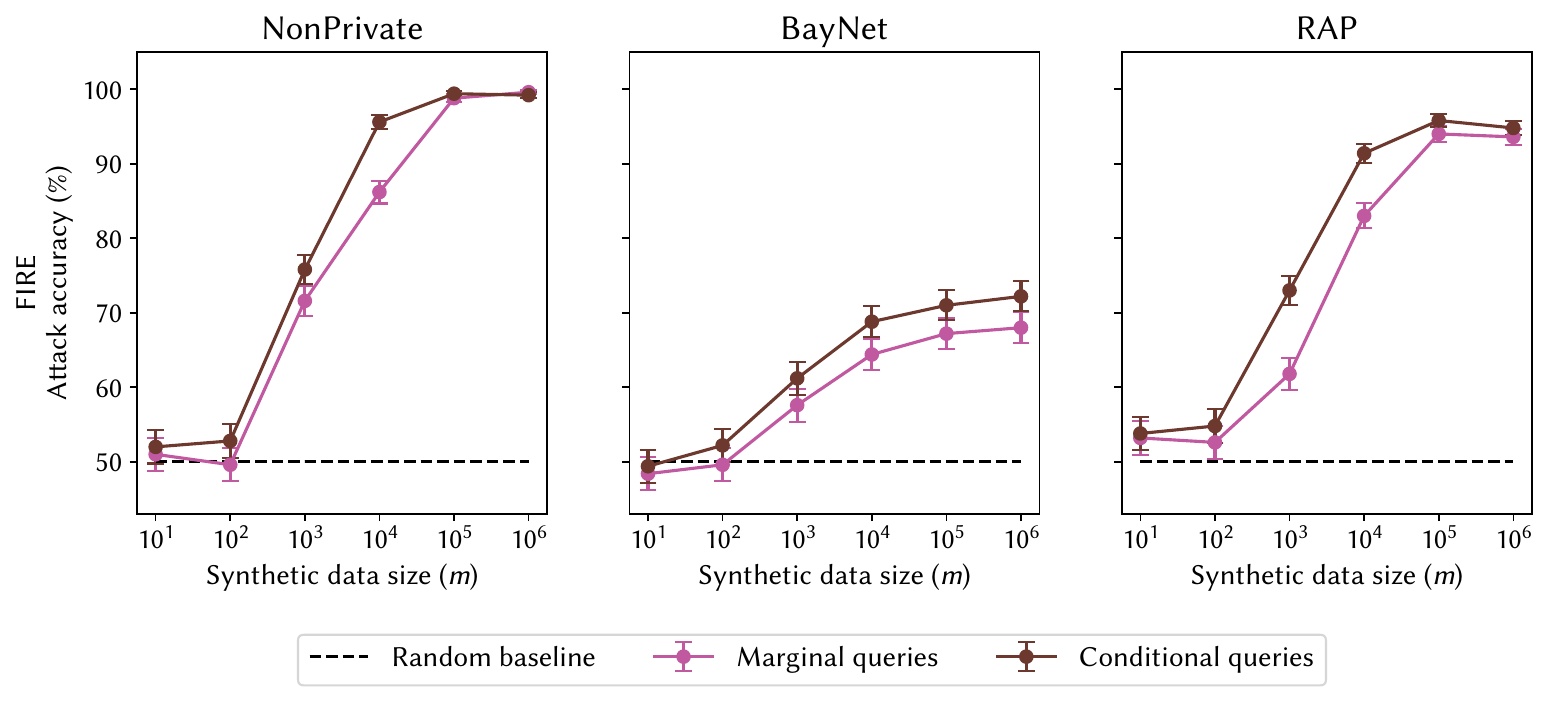}
\centering
\caption{Comparison of attack accuracy of $\Advs$ when using conditional queries compared to marginal queries.}
\label{fig:cond_vs_mar_min}
\end{figure}

\paragraph{Choice of queries.} In our work, we use conditional queries to maximize accuracy, compared to simple marginal queries used for instance by Kasiviswanathan et al.~\cite{kasiviswanathan2013power}. Figure~\ref{fig:cond_vs_mar_min} show that using conditional queries improve the attack accuracy systematically compared to simple marginal queries. For instance, using conditional queries improves the attack accuracy by up to 11.2 p.p. against RAP, at $m = 10^3$ on the FIRE dataset. This suggests that our attack is particularly successful because the conditional queries leverage additional knowledge from quasi-ids available to the \textit{partially informed adversary}.

\begin{figure}[htb]
\includegraphics[width=0.9\linewidth]{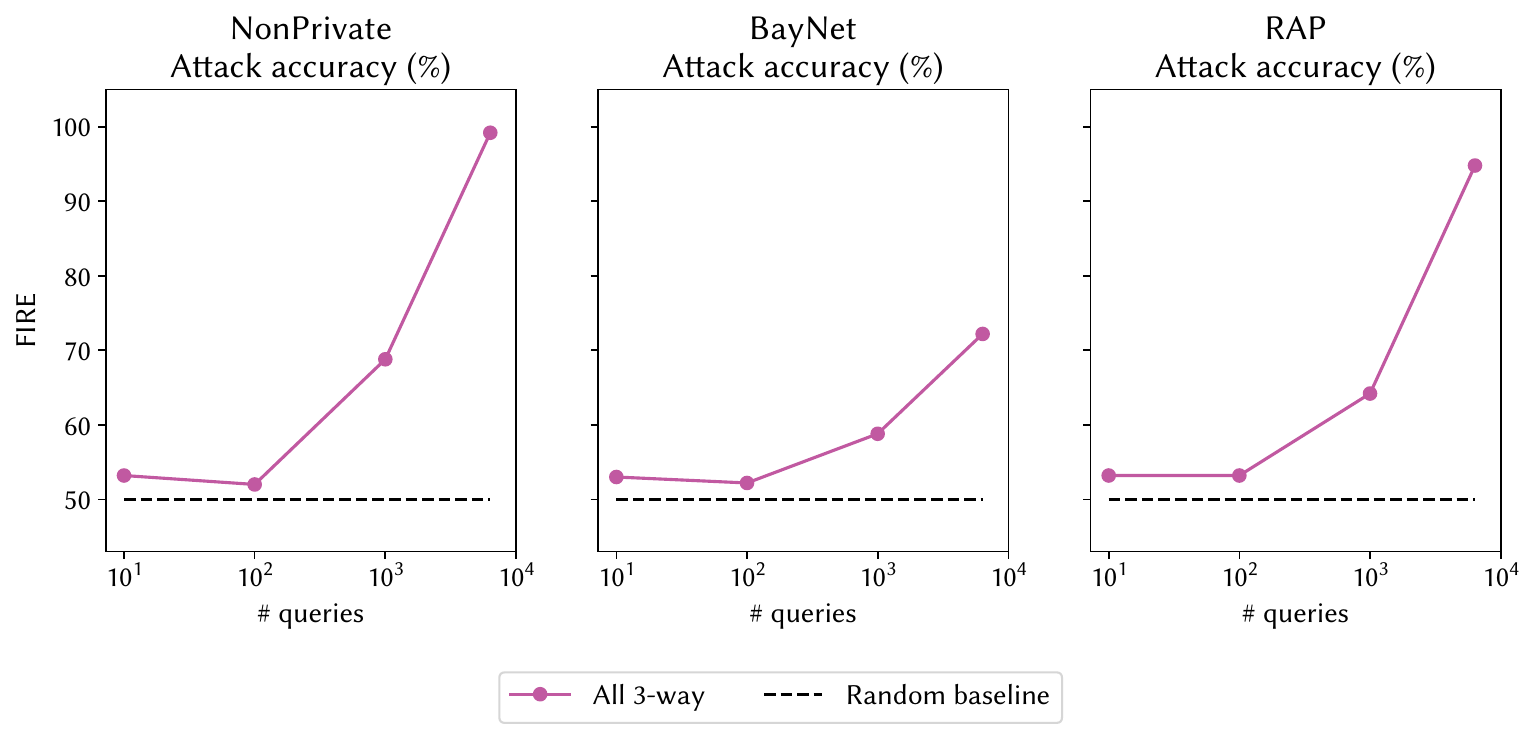}
\centering
\caption{Comparison of attack accuracy of $\Advs$ when uses varying number of queries for synthetic data size, $m = 10^6$}
\label{fig:acc_vs_n_queries_min}
\end{figure}

\paragraph{Number of queries.} Increased data dimensionality and higher computational budget could help an attacker run our attack with more queries. Figure~\ref{fig:acc_vs_n_queries_min} shows that the accuracy of $\Advs$ increases with the number of 3-way conditional queries used. For instance, for the FIRE dataset and the RAP algorithm, increasing the number of queries from $10^3$ to $10^4$ results in a more than 20 p.p. increase in attack accuracy. This suggests that for higher dimensional datasets, our attack could potentially leverage a larger number of available queries to achieve even higher accuracy. Since our attack is no-box, and does not rely on shadow modelling, an adversary in practice will not be computationally constrained by  increased data dimensionality. However, as increased data dimensionality heavily increases the computational cost of generating a large number of synthetic datasets necessary for the evaluation of privacy risk, in our work, we restrict ourselves to more tractable datasets such as ACS and FIRE.

\subsection{Computational cost}
\label{sec:comp_cost}
\begin{table}[H]
    \begin{center}
        \setlength\tabcolsep{2pt} %
        \begin{tabular}{c|ccccc}
            \toprule
            \multirow{2}{*}{$m$} & \multicolumn{5}{c}{No. of queries $|Q|$}\\
            & $10^1$ & $10^2$ & $10^3$ & $10^4$ & $10^5$ \\
            \midrule
            $10^1$ & 2.10 & 2.20 & 2.39 & 4.43 & 25.2 \\
            $10^2$ & 2.10 & 2.22 & 2.39 & 4.47 & 25.6 \\
            $10^3$ & 2.11 & 2.21 & 2.42 & 4.74 & 28.6 \\
            $10^4$ & 2.11 & 2.22 & 2.50 & 5.80 & 39.0 \\
            $10^5$ & 2.19 & 2.41 & 3.75 & 15.4 & 134 \\ 
            $10^6$ & 2.82 & 4.12 & 16.5 & 113 & 1130 \\ 
            \bottomrule
        \end{tabular}
    \end{center}
    \caption{Time taken (s) for single attack for various number of queries and synthetic data sizes, $m$}
    \label{table:comp_cost}
\end{table}
Lastly, in order to show that our attack is in fact practical, we evaluate the computational cost of running our attack with respect to the two main factors---number of queries and synthetic data size. While linear reconstruction attacks have already been known to be practical, most prior work~\cite{cohen2018linear,dwork2007price} have only implemented such attacks with a couple of thousand queries. In our work, we make use of parallelization and efficient commercial solvers such as Gurobi to run our attack efficiently even when tens and hundreds of thousands of queries are used.

To that end, we record the time taken for our attack to run against the NonPrivate algorithm (without loss of generality) on the FIRE dataset for 1 repetition. We vary the number of synthetic data sizes and numbers of queries and report the  cost of running a single attack in Table~\ref{table:comp_cost}. We find that our attack runs very efficiently, taking less than 5 mins to run in almost all settings and finishing in under half an hour even in the most computationally intensive setting (No. of queries = $10^5$, $m = 10^6$). In this setting, the amount of memory required to run a single attack is also small (3.26 GB). Therefore, we can see that our attack runs very efficiently and increasing the number of queries when available is still possible.
\section{Discussion}
\label{sec:discussion}
\subsection{Linear reconstruction attacks}
There has been a rich literature on linear reconstruction attacks (which led to the formulation of \textit{The Fundamental Law of Information Recovery}~\cite{dwork2014algorithmic}), which has thus far not been considered in the context of synthetic data. Informally, the law states that answering too many queries with too much accuracy destroys privacy ``in a spectacular way" and is formally defined in relation to linear reconstruction attacks \cite{dinur2003revealing, dwork2007price, kasiviswanathan2013power}. Linear reconstruction attacks provide theoretical asymptotic results on the amount of secret information that can be reconstructed when a large fraction of linear queries are answered with insufficient noise by any mechanism.

Linear reconstruction attacks are typically studied in the context of systems (e.g. interactive data query systems) that compute some aggregate statistics and add some noise before releasing them (e.g. Gaussian noise subject to clipping and memoization~\cite{cohen2018linear}). In this work, we see synthetic data as a noise-addition mechanism as well---with the noise being the difference between the aggregate statistic computed on the original data and the same statistic computed on the synthetic data. However, unlike systems that add noise in a coherent way, the noise added by most SDG algorithms does not follow a coherent structure. This work therefore shows that it is in fact possible to extend linear reconstruction attacks to synthetic data in practical settings.

\subsection{Attribute Inference Privacy Game}
\label{sec:discussion-privacy-game}
Yeom et al.~formalized the risk of attribute inference as an \textit{attribute inference privacy game}~\cite{yeom2018privacy}. However, Jayaraman et al.~noted that, under this formulation, some of the state of the art attacks can be trivially outperformed by a naive baseline~\cite{jayaraman2022attribute}. They showed that a naive baseline---outputting the most common value as the secret attribute---outperforms three of the previous state of the art attacks. Furthermore, Houssiau et al.~noted that this formulation suffers from the base-rate problem~\cite{houssiau2022tapas}. If the secret attribute, say income, is strongly correlated with non-secret attributes such as age, education, and country, then the secret attribute can be predicted with very high confidence by simply having access to population-level knowledge, even without access to the released model.

In our game, the challenger randomizes the target record’s secret attribute in Step 2. This achieves two goals. Firstly, correlations between the secret and non-secret attributes are broken specifically for the target record. This means that any adversary without access to the synthetic data cannot possibly predict the secret attribute more accurately than the random baseline which solves the base-rate problem that was discussed by Houssiau et al~\cite{houssiau2022tapas}. Secondly, the naive baseline presented by Jayaraman et al.~\cite{jayaraman2022attribute}---predicting the most common secret attribute---cannot possibly perform better than the random baseline.

\subsection{Broader impacts of Attribute Inference Attacks}
Attribute inference attacks are widely considered a significant privacy breach in privacy-preserving data publishing. Indeed, inferring an individual's attribute may pose concrete risks to that person~\cite{workingparty,ico}. The risks posed by attribute inference attacks became well recognized in 1997, when Sweeney demonstrated how to infer medical records relating to Massachusetts governor by linking his quasi-identifiers (sex, date of birth, ZIP code) to his records in a pseudonymized public health dataset~\cite{sweeney2000simple}. However, when the target dataset consists of aggregate or synthetic data, it is not always easy to establish whether a successful attribute inference attack reveals a breach of individual privacy~\cite{mcsherrystat,bun2021statistical}. In fact, simply revealing population-level information, such as correlations, can be sufficient to accurately infer an attribute from quasi-identifiers~\cite{mcsherrystat}.

In our work, we build our attribute inference privacy game to separate out the individual-level information leakage from the population-level inferences discussed by Jayaraman et al.~\cite{jayaraman2022attribute}. Our findings show that $\Advs$ is in fact able to exploit access to the synthetic data beyond what is theoretically possible from population-level knowledge and precisely determine the secret attribute of a single individual. This leakage comes directly from the individual participation in the data, individuals whose secret attribute would not have been otherwise leaked from ``distributional inference''. We believe that our work makes a strong case for attribute inference attacks being considered a significant risk for synthetic data in practice. 

\subsection{Partially informed adversary}
\label{sec:discussion-partially-informed}
In this work, we consider a partially informed adversary: an adversary with access to all the quasi-identifiers $X$ (i.e. the non-secret attributes) of the original dataset. This adversary was previously discussed by Kasiviswanathan et al.~\cite{kasiviswanathan2013power} for medical and retail data. While this is a strong assumption, it is still a much weaker and more practical assumption than the exact knowledge setup used to verify differential privacy guarantees~\cite{jagielski2020auditing}. Moreover, the partially informed setting can be realistic in the context of multiple data releases. For example, national statistics, public health, and medical institution often publish multiple releases from the same original dataset(s)~\cite{xia2023managing,benedetto2018creation}. Some releases will require a particular disclosure method, e.g. de-identification or pseudonymised microdata for low-risk data, while others will rely on synthetic data generation. This is for instance the case for the Survey of Income and Program Participation (SIPP), released both as microdata and as synthetic data (combined with more sensitive IRS tax data)~\cite{benedetto2018creation}.

\subsection{Higher order marginal queries}
\label{sec:discussion-higher-order-marginals}
In this work, our adversary uses 3-way marginal queries to carry out attribute inference attacks. While higher order marginal queries could also be used to mount our attack, this does not naturally lead to stronger attacks. Including higher order marginal queries (such as 4 and 5-way marginal queries) increases the overall number of queries available to the attack thus increasing the amount of knowledge that can potentially be exploited by our reconstruction attack. At the same time, higher order marginal queries have larger amounts of error, even in the NonPrivate setting, thus reducing the overall power of the attack. In \ifusenixfinal the extended version of the paper~\cite{annamalai2023linear}\else Appendix~\ref{appsec:varying_k}\fi, we empirically show that 3-way queries provide the highest accuracy. There are too few 2-way queries involving the secret attribute that could be used to mount a successful attack. On the other hand, including a high number of 4-way queries does not improve the overall performance of our attack.

Recent work by Crețu et al.~\cite{cretu2022querysnout} suggests that machine learning techniques could in fact be adapted to identify informative higher order queries. In the context of query-based systems, they show that evolutionary algorithms can be used to identify a small set of queries that reveal the sensitive attribute of a target record. By viewing SDG algorithms as noisy query-based systems, their attack can potentially be adapted to similarly select a small set of $k$-way marginal queries that are highly informative and can be used by our attack to boost the attack accuracy. As this may require significant research to adapt their technique on synthetic data generation algorithms, we leave this for future work.

\subsection{Binary secret attributes}
Attribute inference attacks are often evaluated on binary attributes. Successfully attacking a binary attribute is sufficient to show that a mechanism is vulnerable to privacy leakage~\cite{cretu2022querysnout,jayaraman2022attribute}.

In theory, it is possible to relax this assumption and modify our attack to target categorical secret attributes. For instance, we can employ one-hot encoding in order to attack a categorical attribute. The attribute $y \in \{0, 1, ..., Y\}$ is encoded into a $Y$-dimensional vector $\mathbf{v} \in \{0,1\}^Y$ where $\mathbf{v}_i = 0\; \forall i \neq Y$ and $\mathbf{v}_y = 1$. Our attack can then be run repeatedly $Y$ times, for each  binary attribute $\mathbf{v}_i$. The $\mathbf{v}_i$ with highest score can be returned as the reconstructed secret attribute $y$, where the score is computed according to an appropriate function on $\mathbf{v} \in \{0,1\}^Y$.

However, successfully mounting this attack against any attribute and testing its accuracy in our evaluation framework remains challenging. Firstly, the modified attack requires $Y$ times more computational power than for a single binary attribute, increasing linearly with the domain of the secret attribute. Furthermore, the error resulting from the $Y$ iterations would accumulate, resulting in a lower attack accuracy. Although this effect could be possibly mitigated by choosing a better score function, addressing these and other challenges is outside the scope of this paper.
\section{Conclusion}
\label{sec:conclusion}
In this work, we studied the privacy-utility tradeoffs of synthetic data generation algorithms for tabular data. We showed how to extend and adapt linear reconstruction attacks for synthetic data in order to successfully perform attribute inference even when the adversary has no access to the underlying generative model. We showed that prior attacks can underestimate the privacy risk of synthetic data by up to 10 percentage points compared to our linear reconstruction attack, suggesting that empirical resistance to attacks requires evaluations on multiple attack methods. Furthermore, our work showed that information leakage on arbitrary records is in fact possible from synthetic data. This suggests that when the synthetic data provides good utility, data subjects can in fact be at risk of privacy leakage.

Our analysis further revealed that generating and releasing a large amount of synthetic records can marginally increase utility while simultaneously enabling much more powerful attacks, a finding that has not been widely studied in prior work. This has important implications when releasing synthetic data: evaluations on small synthetic benchmarks may underestimate the actual risks if larger synthetic datasets are released.

We finally evaluated differentially private synthetic mechanisms. We showed that, while some mechanisms provide worse privacy-utility tradeoffs in practice, recent mechanisms such as $\RAPDP$ can in fact provide better empirical privacy-utility tradeoffs than non-DP mechanisms in specific settings and synthetic data sizes.

In conclusion, our results suggest that current synthetic data generation mechanisms may not simultaneously preserve high utility and high privacy at the same time. Privacy-utility tradeoff analyses are highly context dependent and require careful calibration of privacy and utility thresholds to make definitive conclusions. Therefore, synthetic data on its own should not be seen as an all encompassing solution to privacy-preserving data publishing. We hope that our work will support practitioners to evaluate the robustness of synthetic data against violations of privacy and anonymity.

\paragraph{Acknowledgements.}
We thank our anonymous reviewers and shepherd for providing us with valuable feedback that helped improve our article. We would also like to additionally thank Ana-Maria Creţu, Benjamin Tan Hong Meng, Emiliano De Cristofaro, Florimond Houssiau, and Khin Mi Mi Aung for their valuable support and feedback on earlier versions of this work. The authors acknowledge support from the Royal Society Research Grant RG\textbackslash R2\textbackslash 232035, the John Fell OUP Research Fund, the EPSRC grant EP/W016419/1, the National Science Scholarship (PhD), and the RIE2020 Advanced Manufacturing and Engineering (AME) Programmatic Program (Award A19E3b0099).

\bibliographystyle{plain}
\bibliography{bib}

\begin{thebibliography}{10}

\bibitem{uscensus2020}
US~Census 2020.
\newblock {Data Metrics for 2020 Disclosure Avoidance}.
\newblock
  \url{https://www2.census.gov/programs-surveys/decennial/2020/program-management/data-product-planning/disclosure-avoidance-system/2020-03-25-data-metrics-2020-da.pdf},
  2020.

\bibitem{aydore2021differentially}
Sergul Aydore, William Brown, Michael Kearns, Krishnaram Kenthapadi, Luca
  Melis, Aaron Roth, and Ankit~A Siva.
\newblock {Differentially Private Query Release Through Adaptive Projection}.
\newblock In {\em ICML}, 2021.

\bibitem{bagdasaryan2019differential}
Eugene Bagdasaryan, Omid Poursaeed, and Vitaly Shmatikov.
\newblock {Differential privacy has disparate impact on model accuracy}.
\newblock In {\em Advances in neural information processing systems}, 2019.

\bibitem{benedetto2018creation}
Gary Benedetto, Jordan~C Stanley, Evan Totty, et~al.
\newblock {The creation and use of the SIPP synthetic Beta v7.0}.
\newblock {\em US Census Bureau}, 2018.

\bibitem{dputil2021}
Claire~McKay Bowen.
\newblock {Utility Metrics for Differential Privacy: No One-Size-Fits-All}.
\newblock
  \url{https://www.nist.gov/blogs/cybersecurity-insights/utility-metrics-differential-privacy-no-one-size-fits-all},
  2021.

\bibitem{bun2021statistical}
Mark Bun, Damien Desfontaines, Cynthia Dwork, Moni Naor, Kobbi Nissim, Aaron
  Roth, Adam Smith, Thomas Steinke, Jonathan Ullman, and Salil Vadhan.
\newblock Statistical inference is not a privacy violation.
\newblock
  \url{https://differentialprivacy.org/inference-is-not-a-privacy-violation/},
  2021.

\bibitem{cai2021data}
Kuntai Cai, Xiaoyu Lei, Jianxin Wei, and Xiaokui Xiao.
\newblock {Data Synthesis via Differentially Private Markov Random Fields}.
\newblock {\em VLDB Endowment}, 14(11), 2021.

\bibitem{carlini2022membership}
Nicholas Carlini, Steve Chien, Milad Nasr, Shuang Song, Andreas Terzis, and
  Florian Tramer.
\newblock {Membership Inference Attacks From First Principles}.
\newblock In {\em IEEE S\&P}, 2022.

\bibitem{carlini2021extracting}
Nicholas Carlini, Florian Tramer, Eric Wallace, Matthew Jagielski, Ariel
  Herbert-Voss, Katherine Lee, Adam Roberts, Tom Brown, Dawn Song, Ulfar
  Erlingsson, et~al.
\newblock {Extracting Training Data from Large Language Models}.
\newblock In {\em USENIX}, 2021.

\bibitem{chen2020gan}
Dingfan Chen, Ning Yu, Yang Zhang, and Mario Fritz.
\newblock {GAN-Leaks: A Taxonomy of Membership Inference Attacks against
  Generative Models}.
\newblock In {\em ACM CCS}, 2020.

\bibitem{cohen2022attacks}
Aloni Cohen.
\newblock {Attacks on Deidentification's Defenses}.
\newblock In {\em USENIX}, 2022.

\bibitem{cohen2018linear}
Aloni Cohen and Kobbi Nissim.
\newblock {Linear Program Reconstruction in Practice}.
\newblock {\em Journal of Privacy and Confidentiality}, 10(1), 2020.

\bibitem{cretu2022querysnout}
Ana-Maria Cretu, Florimond Houssiau, Antoine Cully, and Yves-Alexandre
  de~Montjoye.
\newblock {QuerySnout: Automating the Discovery of Attribute Inference Attacks
  against Query-Based Systems}.
\newblock In {\em ACM CCS}, 2022.

\bibitem{cummings2023challenges}
Rachel Cummings, Damien Desfontaines, David Evans, Roxana Geambasu, Matthew
  Jagielski, Yangsibo Huang, Peter Kairouz, Gautam Kamath, Sewoong Oh, Olga
  Ohrimenko, et~al.
\newblock Challenges towards the next frontier in privacy.
\newblock {\em arXiv:2304.06929}, 2023.

\bibitem{fire2022}
{DataSF}.
\newblock {Fire Department Calls for Service}.
\newblock
  \url{https://data.sfgov.org/Public-Safety/Fire-Department-Calls-for-Service/nuek-vuh3},
  2020.

\bibitem{deming1940least}
W~Edwards Deming and Frederick~F Stephan.
\newblock {On a Least Squares Adjustment of a Sampled Frequency Table When the
  Expected Marginal Totals are Known}.
\newblock {\em The Annals of Mathematical Statistics}, 11(4), 1940.

\bibitem{dick2022confidence}
Travis Dick, Cynthia Dwork, Michael Kearns, Terrance Liu, Aaron Roth, Giuseppe
  Vietri, and Zhiwei~Steven Wu.
\newblock {Confidence-ranked reconstruction of census microdata from published
  statistics}.
\newblock {\em PNAS}, 120(8), 2023.

\bibitem{ding2021retiring}
Frances Ding, Moritz Hardt, John Miller, and Ludwig Schmidt.
\newblock {Retiring Adult: New Datasets for Fair Machine Learning}.
\newblock In {\em NeurIPS}, 2021.

\bibitem{dinur2003revealing}
Irit Dinur and Kobbi Nissim.
\newblock {Revealing Information while Preserving Privacy}.
\newblock In {\em ACM Symposium on Principles of Database Systems}, 2003.

\bibitem{dwork2006calibrating}
Cynthia Dwork, Frank McSherry, Kobbi Nissim, and Adam Smith.
\newblock {Calibrating Noise to Sensitivity in Private Data Analysis}.
\newblock In {\em Theory of Cryptography}, 2006.

\bibitem{dwork2007price}
Cynthia Dwork, Frank McSherry, and Kunal Talwar.
\newblock {The price of privacy and the limits of LP decoding}.
\newblock In {\em ACM Symposium on Theory of Computing}, 2007.

\bibitem{dwork2014algorithmic}
Cynthia Dwork and Aaron Roth.
\newblock The algorithmic foundations of differential privacy.
\newblock {\em Foundations and Trends in Theoretical Computer Science},
  9(3--4), 2014.

\bibitem{fredrikson2014privacy}
Matthew Fredrikson, Eric Lantz, Somesh Jha, Simon Lin, David Page, and Thomas
  Ristenpart.
\newblock {Privacy in Pharmacogenetics: An End-to-End Case Study of
  Personalized Warfarin Dosing}.
\newblock In {\em USENIX}, 2014.

\bibitem{gadotti2022pool}
Andrea Gadotti, Florimond Houssiau, Meenatchi Sundaram Muthu~Selva Annamalai,
  and Yves-Alexandre de~Montjoye.
\newblock {Pool Inference Attacks on Local Differential Privacy: Quantifying
  the Privacy Guarantees of Apple's Count Mean Sketch in Practice}.
\newblock In {\em USENIX}, 2022.

\bibitem{gadotti2019signal}
Andrea Gadotti, Florimond Houssiau, Luc Rocher, Benjamin Livshits, and
  Yves-Alexandre de~Montjoye.
\newblock {When the signal is in the noise: Exploiting Diffix's Sticky Noise}.
\newblock In {\em USENIX}, 2019.

\bibitem{ganev2022robin}
Georgi Ganev, Bristena Oprisanu, and Emiliano De~Cristofaro.
\newblock {Robin Hood and Matthew Effects: Differential Privacy Has Disparate
  Impact on Synthetic Data}.
\newblock In {\em ICML}, 2022.

\bibitem{giomi2022unified}
Matteo Giomi, Franziska Boenisch, Christoph Wehmeyer, and Borb{\'a}la
  Tasn{\'a}di.
\newblock {A Unified Framework for Quantifying Privacy Risk in Synthetic Data}.
\newblock In {\em PETS}, 2023.

\bibitem{hayes2017logan}
Jamie Hayes, Luca Melis, George Danezis, and Emiliano De~Cristofaro.
\newblock {LOGAN: Membership Inference Attacks Against Generative Models}.
\newblock In {\em PETS}, 2019.

\bibitem{hilprecht2019monte}
Benjamin Hilprecht, Martin H{\"a}rterich, and Daniel Bernau.
\newblock {Monte Carlo and Reconstruction Membership Inference Attacks against
  Generative Models}.
\newblock In {\em PETS}, 2019.

\bibitem{houssiau2022tapas}
Florimond Houssiau, James Jordon, Samuel~N Cohen, Andrew Elliott, James Geddes,
  Callum Mole, Camila Rangel-Smith, and Lukasz Szpruch.
\newblock {TAPAS: a Toolbox for Adversarial Privacy Auditing of Synthetic
  Data}.
\newblock In {\em NeurIPS 2022 Workshop on Synthetic Data for Empowering ML
  Research}, 2022.

\bibitem{hradec2022multipurpose}
Jiri Hradec, Massimo Craglia, Margherita Di~Leo, Sarah De~Nigris, Nicole
  Ostlaender, and Nicholas Nicholson.
\newblock {Multipurpose synthetic population for policy applications}.
\newblock {\em JRC128595}, 2022.

\bibitem{jagielski2020auditing}
Matthew Jagielski, Jonathan Ullman, and Alina Oprea.
\newblock {Auditing Differentially Private Machine Learning: How Private is
  Private SGD?}
\newblock In {\em NeurIPS}, 2020.

\bibitem{jayaraman2022attribute}
Bargav Jayaraman and David Evans.
\newblock {Are Attribute Inference Attacks Just Imputation?}
\newblock In {\em ACM CCS}, 2022.

\bibitem{jeong2016copula}
Byungduk Jeong, Wonjoon Lee, Deok-Soo Kim, and Hayong Shin.
\newblock {Copula-Based Approach to Synthetic Population Generation}.
\newblock {\em PloS One}, 11(8), 2016.

\bibitem{kasiviswanathan2013power}
Shiva~Prasad Kasiviswanathan, Mark Rudelson, and Adam Smith.
\newblock {The Power of Linear Reconstruction Attacks}.
\newblock In {\em ACM-SIAM Symposium on Discrete Algorithms}, 2013.

\bibitem{kenny2021use}
Christopher~T Kenny, Shiro Kuriwaki, Cory McCartan, Evan~TR Rosenman, Tyler
  Simko, and Kosuke Imai.
\newblock {The use of differential privacy for census data and its impact on
  redistricting: The case of the 2020 US Census}.
\newblock {\em Science Advances}, 7(41), 2021.

\bibitem{komarova2015estimation}
Tatiana Komarova, Denis Nekipelov, and Evgeny Yakovlev.
\newblock {Estimation of treatment effects from combined data: Identification
  versus data security}.
\newblock In {\em Economic Analysis of the Digital Economy}, pages 279--308.
  University of Chicago Press, April 2015.

\bibitem{liu2021iterative}
Terrance Liu, Giuseppe Vietri, and Steven~Z Wu.
\newblock {Iterative Methods for Private Synthetic Data: Unifying Framework and
  New Methods}.
\newblock In {\em NeurIPS}, 2021.

\bibitem{loken2017measurement}
Eric Loken and Andrew Gelman.
\newblock {Measurement error and the replication crisis}.
\newblock {\em Science}, 355(6325), 2017.

\bibitem{lu2019empirical}
Pei-Hsuan Lu, Pang-Chieh Wang, and Chia-Mu Yu.
\newblock {Empirical Evaluation on Synthetic Data Generation with Generative
  Adversarial Network}.
\newblock In {\em International Conference on Web Intelligence, Mining and
  Semantics}, 2019.

\bibitem{mckenna2021winning}
Ryan McKenna, Gerome Miklau, and Daniel Sheldon.
\newblock Winning the nist contest: A scalable and general approach to
  differentially private synthetic data.
\newblock {\em Journal of Privacy and Confidentiality}, 11(3), 2021.

\bibitem{mckenna2022aim}
Ryan McKenna, Brett Mullins, Daniel Sheldon, and Gerome Miklau.
\newblock {AIM: An Adaptive and Iterative Mechanism for Differentially Private
  Synthetic Data}.
\newblock {\em VLDB Endowment}, 15(11), 2022.

\bibitem{mcsherrystat}
Frank McSherry.
\newblock {Statistical inference considered harmful}.
\newblock
  \url{https://github.com/frankmcsherry/blog/blob/master/posts/2016-06-14.md},
  2016.

\bibitem{nasr2019comprehensive}
Milad Nasr, Reza Shokri, and Amir Houmansadr.
\newblock {Comprehensive Privacy Analysis of Deep Learning: Passive and Active
  White-box Inference Attacks against Centralized and Federated Learning}.
\newblock In {\em IEEE S\&P}, 2019.

\bibitem{tvdcensus2020}
National~Academies of~Sciences, Engineering, and Medicine.
\newblock {\em {2020 Census Data Products: Data Needs and Privacy
  Considerations: Proceedings of a Workshop}}.
\newblock National Academies Press, 2020.

\bibitem{dpsynth2018}
National~Institute of~Standards and Technology.
\newblock {2018 Differential Privacy Synthetic Data Challenge}.
\newblock
  \url{https://www.nist.gov/ctl/pscr/open-innovation-prize-challenges/past-prize-challenges/2018-differential-privacy-synthetic},
  2018.

\bibitem{workingparty}
Information~Commissioner's Office.
\newblock {Article 29 Data Protection Working Party. Opinion 05/2014 on
  anonymisation techniques.}
\newblock
  \url{https://ec.europa.eu/justice/article-29/documentation/opinion-recommendation/files/2014/wp216_en.pdf},
  2014.

\bibitem{ico}
Information~Commissioner's Office.
\newblock {Chapter 2: How do we ensure anonymisation is effective?}
\newblock
  \url{https://ico.org.uk/media/about-the-ico/documents/4018606/chapter-2-anonymisation-draft.pdf},
  2021.

\bibitem{ohm2009broken}
Paul Ohm.
\newblock {Broken Promises of Privacy: Responding to the Surprising Failure of
  Anonymization}.
\newblock {\em UCLA Law Review}, 57, 2009.

\bibitem{oprisanu2021utility}
Bristena Oprisanu, Georgi Ganev, and Emiliano De~Cristofaro.
\newblock {On Utility and Privacy in Synthetic Genomic Data}.
\newblock In {\em NDSS}, 2022.

\bibitem{ping2017datasynthesizer}
Haoyue Ping, Julia Stoyanovich, and Bill Howe.
\newblock {DataSynthesizer: Privacy-Preserving Synthetic Datasets}.
\newblock In {\em International Conference on Scientific and Statistical
  Database Management}, 2017.

\bibitem{pyrgelis2017what}
Apostolos Pyrgelis, Carmela Troncoso, and Emiliano~De Cristofaro.
\newblock {What Does The Crowd Say About You? Evaluating Aggregation-based
  Location Privacy}.
\newblock In {\em PETS}, 2017.

\bibitem{pyrgelis2018knock}
Apostolos Pyrgelis, Carmela Troncoso, and Emiliano~De Cristofaro.
\newblock {Knock Knock, Who's There? Membership Inference on Aggregate Location
  Data}.
\newblock In {\em NDSS}, 2018.

\bibitem{reiter2005using}
Jerome~P Reiter.
\newblock {Using CART to Generate Partially Synthetic Public Use Microdata}.
\newblock {\em Journal of Official Statistics}, 21(3), 2005.

\bibitem{ridgeway2021challenge}
Diane Ridgeway, Mary~F Theofanos, Terese~W Manley, and Christine Task.
\newblock {Challenge Design and Lessons Learned from the 2018 Differential
  Privacy Challenges}, 2021.

\bibitem{rocher2019estimating}
Luc Rocher, Julien~M Hendrickx, and Yves-Alexandre de~Montjoye.
\newblock {Estimating the success of re-identifications in incomplete datasets
  using generative models}.
\newblock {\em Nature Communications}, 10(1), 2019.

\bibitem{rosenblatt2020differentially}
Lucas Rosenblatt, Xiaoyan Liu, Samira Pouyanfar, Eduardo de~Leon, Anuj Desai,
  and Joshua Allen.
\newblock {Differentially Private Synthetic Data: Applied Evaluations and
  Enhancements}.
\newblock {\em arXiv:2011.05537}, 2020.

\bibitem{rubin1993statistical}
Donald~B Rubin.
\newblock {Statistical Disclosure Limitation}.
\newblock {\em Journal of official Statistics}, 9(2), 1993.

\bibitem{shokri2017membership}
Reza Shokri, Marco Stronati, Congzheng Song, and Vitaly Shmatikov.
\newblock {Membership Inference Attacks against Machine Learning Models}.
\newblock In {\em IEEE S\&P}, 2017.

\bibitem{royalsocietysynth}
The~Royal Society.
\newblock {What is synthetic data, and how can it advance research and
  development?}
\newblock \url{https://royalsociety.org/blog/2022/05/synthetic-data/}, 2022.

\bibitem{stadler2022synthetic}
Theresa Stadler, Bristena Oprisanu, and Carmela Troncoso.
\newblock {Synthetic Data -- Anonymisation Groundhog Day}.
\newblock In {\em USENIX}, 2022.

\bibitem{sweeney1997weaving}
Latanya Sweeney.
\newblock {Weaving Technology and Policy Together to Maintain Confidentiality}.
\newblock {\em The Journal of Law, Medicine \& Ethics}, 25(2-3), 1997.

\bibitem{sweeney2000simple}
Latanya Sweeney.
\newblock Simple demographics often identify people uniquely.
\newblock {\em Health (San Francisco)}, 2000.

\bibitem{tao2021benchmarking}
Yuchao Tao, Ryan McKenna, Michael Hay, Ashwin Machanavajjhala, and Gerome
  Miklau.
\newblock {Benchmarking Differentially Private Synthetic Data Generation
  Algorithms}.
\newblock {\em arXiv:2112.09238}, 2021.

\bibitem{tramer2021differentially}
Florian Tramer and Dan Boneh.
\newblock {Differentially Private Learning Needs Better Features (or Much More
  Data)}.
\newblock In {\em International Conference on Learning Representations}, 2021.

\bibitem{Wagner2018-me}
Isabel Wagner and David Eckhoff.
\newblock {Technical Privacy Metrics: A Systematic Survey}.
\newblock {\em ACM Computing Surveys}, 51(3), 2018.

\bibitem{xia2023managing}
Weiyi Xia, Melissa Basford, Robert Carroll, Ellen~Wright Clayton, Paul Harris,
  Murat Kantacioglu, Yongtai Liu, Steve Nyemba, Yevgeniy Vorobeychik, Zhiyu
  Wan, and Bradley~A Malin.
\newblock {Managing re-identification risks while providing access to the All
  of Us research program}.
\newblock {\em Journal of the American Medical Informatics Association}, 30(5),
  2023.

\bibitem{xu2019modeling}
Lei Xu, Maria Skoularidou, Alfredo Cuesta-Infante, and Kalyan Veeramachaneni.
\newblock {Modeling Tabular data using Conditional GAN}.
\newblock In {\em NeurIPS}, 2019.

\bibitem{yale2019assessing}
Andrew Yale, Saloni Dash, Ritik Dutta, Isabelle Guyon, Adrien Pavao, and
  Kristin~P Bennett.
\newblock {Assessing privacy and quality of synthetic health data}.
\newblock In {\em ACM Artificial Intelligence for Data Discovery and Reuse},
  2019.

\bibitem{ye2022enhanced}
Jiayuan Ye, Aadyaa Maddi, Sasi~Kumar Murakonda, Vincent Bindschaedler, and Reza
  Shokri.
\newblock {Enhanced Membership Inference Attacks against Machine Learning
  Models}.
\newblock In {\em ACM CCS}, 2022.

\bibitem{yeom2018privacy}
Samuel Yeom, Irene Giacomelli, Matt Fredrikson, and Somesh Jha.
\newblock {Privacy Risk in Machine Learning: Analyzing the Connection to
  Overfitting}.
\newblock In {\em IEEE Computer Security Foundations Symposium}, 2018.

\bibitem{zhang2017privbayes}
Jun Zhang, Graham Cormode, Cecilia~M Procopiuc, Divesh Srivastava, and Xiaokui
  Xiao.
\newblock {PrivBayes: Private Data Release via Bayesian Networks}.
\newblock {\em ACM Transactions on Database Systems}, 42(4), 2017.

\bibitem{zhang2022membership}
Ziqi Zhang, Chao Yan, and Bradley~A Malin.
\newblock {Membership inference attacks against synthetic health data}.
\newblock {\em Journal of Biomedical Informatics}, 125, 2022.

\bibitem{zhu2023data}
Derui Zhu, Dingfan Chen, Jens Grossklags, and Mario Fritz.
\newblock {Data Forensics in Diffusion Models: A Systematic Analysis of
  Membership Privacy}.
\newblock {\em arXiv:2302.07801}, 2023.

\end{thebibliography}

\appendix
\section{Descriptions of synthetic data generation algorithms}
\label{appsec:sdg_algos}
\hspace{\parindent} \textit{BayNet}~\cite{zhang2017privbayes}. Bayesian networks are trained by building a probabilistic graphical model that represents the conditional dependence between attributes in the original dataset. This graphical model factorizes the joint distribution between the attributes into a product of low dimensional conditional distributions. The algorithm is defined by the degree of the network which is a parameter that defines the number of attributes in the conditional distributions. We use Ping et al.'s implementation in the DataSynthesizer \cite{ping2017datasynthesizer} library and additionally, set the degree of the network to be 3 in order for 3-way marginal queries to be well preserved in our utility analysis. \\

\textit{PrivBayes}~\cite{zhang2017privbayes} is the differentially private version of the BayNet algorithm. Differentially private laplace noise is added to the conditional distributions such that the overall algorithm satisfies $\varepsilon$-DP. We use Ping et al.'s implementation in the DataSynthesizer~\cite{ping2017datasynthesizer} library and set the degree of the network to 3 similar to the BayNet algorithm. \\

\textit{RAP }\cite{aydore2021differentially}. The relaxed Adaptive Projection algorithm follows the `select-measure-generate' approach. Firstly, a set of summary statistics of interest are selected. Secondly, these statistics are measured on the original dataset. Finally, an optimisation model is used to generate a synthetic dataset with consistent summary statistics. Specifically, the RAP algorithm takes as input a set of differentiable functions $\mathbf{Q}$ and outputs a synthetic dataset using a iterative optimization algorithm (such as the stochastic gradient descent):
\begin{equation*}
    \argmin_\mathcal{S} || \mathbf{Q}(\mathcal{S}) - \mathbf{Q}(\mathcal{D}) ||_2
\end{equation*}
We set the number of optimization iterations to 2000 and set $\mathbf{Q}$ to be the continuous relaxation of all 3-way marginal queries as 3-way marginal queries are important statistics that the original authors intended to be preserved by the algorithm. As such, this makes RAP particularly vulnerable to linear reconstruction attacks on marginal queries. \\

\textit{$\RAPDP$}~\cite{liu2021iterative} is the differentially private version of the RAP algorithm. Differentially private gaussian noise is added to the query measurements made by RAP ($\mathbf{Q}(\mathcal{D})$). Additionally, the exponential mechanism is used to select queries with the largest query errors in each iteration to be optimized. Lastly, the softmax function is used to normalize the output of the algorithm before being projected back to the data domain. Similar to RAP, we set the number of iterations to 2000 and set $\mathbf{Q}$ to be the continuous relaxation of all 3-way marginal queries. $\delta$ is set to $\frac{1}{n^2} = 10^{-6}$. \\

\textit{CTGAN}~\cite{xu2019modeling} is an extension of the Generative Adversarial Network model that trains two deep neural networks, a generative network and a discriminative network, in tandem to generate synthetic data. The generative network takes as input random values from a gaussian distribution and outputs samples in the domain, $\mathcal{X}$. The generative network is then trained to eventually output samples that are indistinguishable from samples from the original dataset by the discriminative network. Random noise is then provided as input to the generative network that generates synthetic samples that form the synthetic dataset. CTGAN is a variant of this architecture and uses a conditional generative network (a generative network that receives the class label as input) and mode-specific normalization to generate synthetic datasets in the context of tabular data. \\

\textit{IndHist}, for ‘independent histogram’, is a baseline algorithm for synthetic data generation~\cite{ping2017datasynthesizer}, independently sampling from the individual distribution of each attribute. Any association between attributes are removed in the synthetic dataset. In particular, this means the best attribute inference attack simply outputs the most common value of the secret attribute, resulting in an attack accuracy of 0.5. In this setting, the synthetic dataset is expected to have very low utility but have good privacy with respect to attribute inference. We use Ping et al.'s~\cite{ping2017datasynthesizer} implementation in the DataSynthesizer library to evaluate this algorithm.

\ifusenixfinal
\else
    \section{Attack accuracy of \texorpdfstring{$\Advs$}{} for varying \texorpdfstring{$k\text{-way}$}{} queries}
    \label{appsec:varying_k}
    In order to find the optimal marginal queries that can be used by our attack, we run our attack with varying $k\text{-way}$ queries. Figure~\ref{fig:2_vs_3_vs_4way} shows the attack accuracies when different $k\text{-way}$ queries were used by the $\Advs$ attack. Since there were more than 100000 $4\text{-way}$ queries (122000 for ACS and 168000 for FIRE on average across 500 games), we use a random sample of 100000 $4\text{-way}$ queries but note that this is still 1 order of magnitude more than the total number of $3\text{-way}$ marginal queries (5000 on average across the 500 games). The total number of $2\text{-way}$ marginal queries on the other hand is only 100 on average across the 500 games.
    
    We observe that the attack is successful when using $3\text{-way}$ and $4\text{-way}$ queries but not for $2\text{-way}$ queries for the RAP and BayNet algorithms. For the CTGAN and IndHist algorithms, the attack is not successful regardless of the type of queries used. This, we believe is because there are not enough $2\text{-way}$ queries to mount a successful attack. Interestingly, $4\text{-way}$ queries result in good attack accuracies even for the RAP algorithm that specifically minimizes the error in $3\text{-way}$ marginal queries only. This suggests that even though the RAP algorithm is specifically set to minimize $3\text{-way}$ marginal queries, a large number of $4\text{-way}$ queries are preserved as well.
    
    We finally test if combining all three different types of queries improves the overall attack accuracy. We show that while combining all three different types of queries gives very slight improvements over just using the $4\text{-way}$ queries, using just the $3\text{-way}$ queries still performs better than the combination of all queries most likely due to the smaller amount of total noise present in the $3\text{-way}$ queries.
    
    \begin{figure}[H]
    \includegraphics[width=0.6\linewidth]{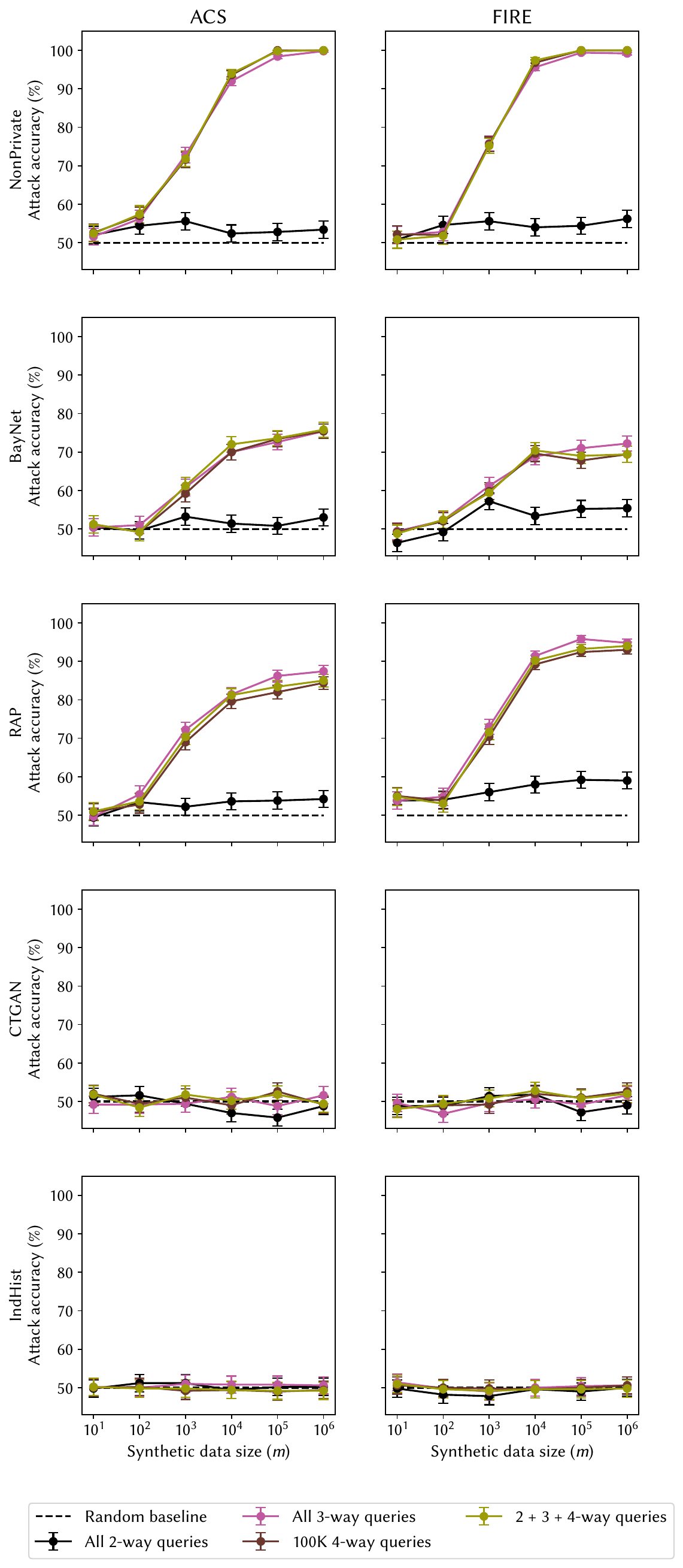}
    \centering
    \caption{Comparison of attack accuracy of $\Advs$ when it uses $2\text{-way}$, $3\text{-way}$, $4\text{-way}$ queries or a combination of the queries. While the maximum possible number of $2\text{-way}$ and $3\text{-way}$ queries are used, 100000 $4\text{-way}$ queries was randomly sampled due to computational limitations}
    \label{fig:2_vs_3_vs_4way}
    \end{figure}

    \section{Attack accuracy of \texorpdfstring{$\Advs$}{} for varying number of queries}
    \label{appsec:varying_n_queries}
    In order to investigate if the attack can perform better when additional queries are available, we run our attack with varying number of queries. Figure~\ref{fig:acc_vs_n_queries} shows the accuracy of $\Advs$ when varying numbers of queries are used for the attack for a synthetic data size of $m = 10^6$. Unsurprisingly, we notice that when there are more queries, the attack performs better. Notably, for the FIRE dataset and the RAP algorithm, increasing the number of queries from $10^3$ to $10^4$ results in a more than 20 p.p. increase in attack accuracy for both the $3\text{-way}$ and $4\text{-way}$ queries. This suggests that for higher dimensional datasets with larger numbers of attributes, our attack will be able to leverage the larger number of available queries to achieve even higher accuracies. However, we note that this is not a linear relationship as we see that past $10^4$ queries, the increase in attack accuracy for the $4\text{-way}$ queries becomes smaller.
    
    \begin{figure}[H]
    \includegraphics[width=0.6\linewidth]{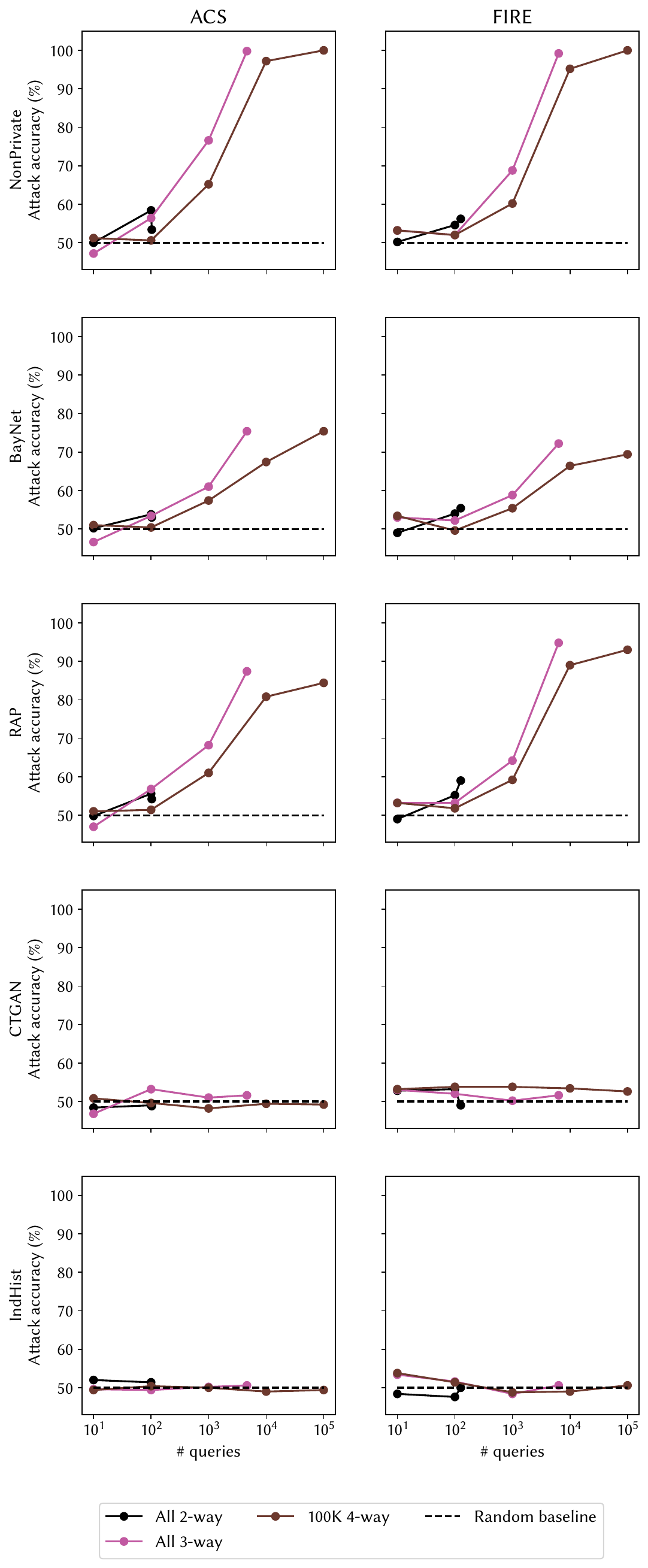}
    \centering
    \caption{Comparison of attack accuracy of $\Advs$ when uses varying number of queries for synthetic data size, $m = 10^6$.}
    \label{fig:acc_vs_n_queries}
    \end{figure}

    \section{Comparison of attack accuracy between $\Advs$, $\Advw$ and $\Advi$}
    \label{appsec:recon_vs_dcr_full}
    Figure~\ref{fig:recon_vs_dcr_full} shows the attack accuracies between our attack, $\Advs$ and the previous state of the art attacks $\Advw$ and $\Advi$ for different synthetic data sizes. We note that while the state of the art attack performs better when the size of the synthetic dataset is smaller ($m = 10^3$, $10^4$), when the size of the synthetic data is larger (and subsequently, of sufficient utility), our attack performs better ($m = 10^5$, $10^6$). 
    \begin{figure}[H]
    \includegraphics[width=\linewidth]{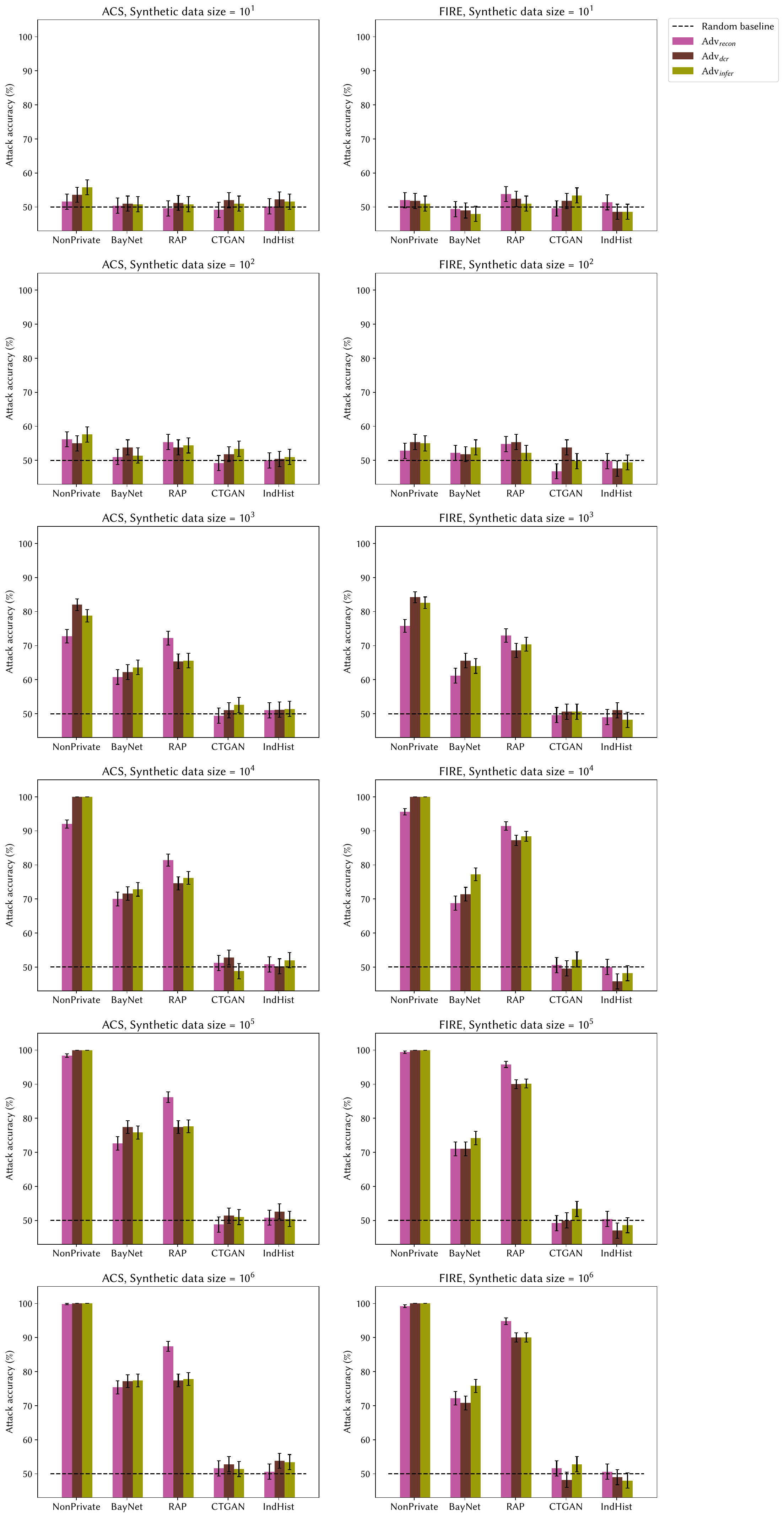}
    \centering
    \caption{Comparison of attack accuracy between $\Advs$ and previous state of the art $\Advw$ for synthetic data sizes $m = 10^1$ to $10^6$}
    \label{fig:recon_vs_dcr_full}
    \end{figure}
    
    \section{Privacy-utility tradeoffs for $\Advs$, $\Advw$, and $\Advi$}
    \label{appsec:priv_util_tradeoffs}
    
    \begin{figure}[H]
    \includegraphics[width=\linewidth]{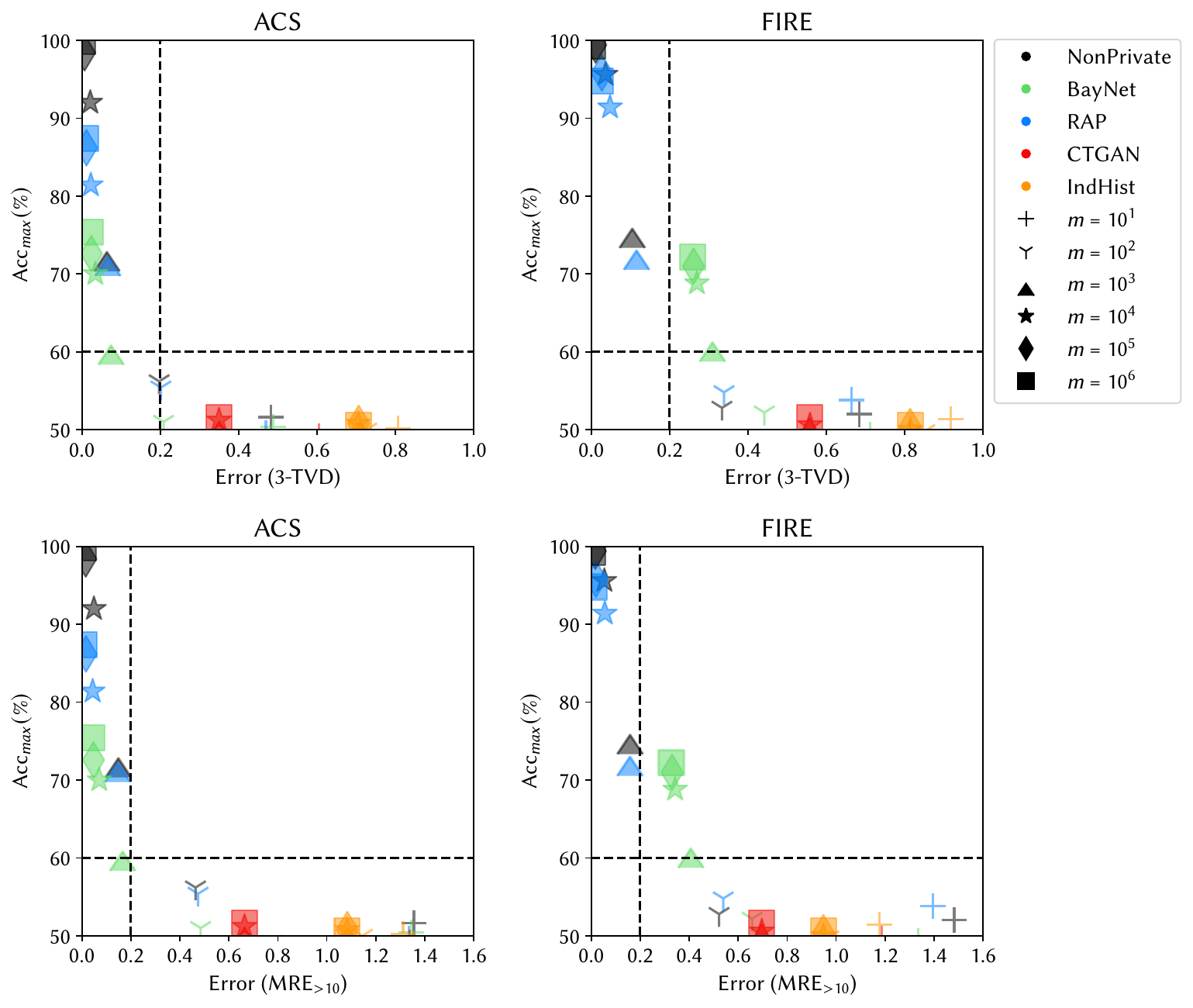}
    \caption{Tradeoff between privacy and utility for $\Advs$}
    \centering
    \label{fig:priv_utility_tradeoff_recon}
    \end{figure}
    
    \begin{figure}[H]
    \includegraphics[width=\linewidth]{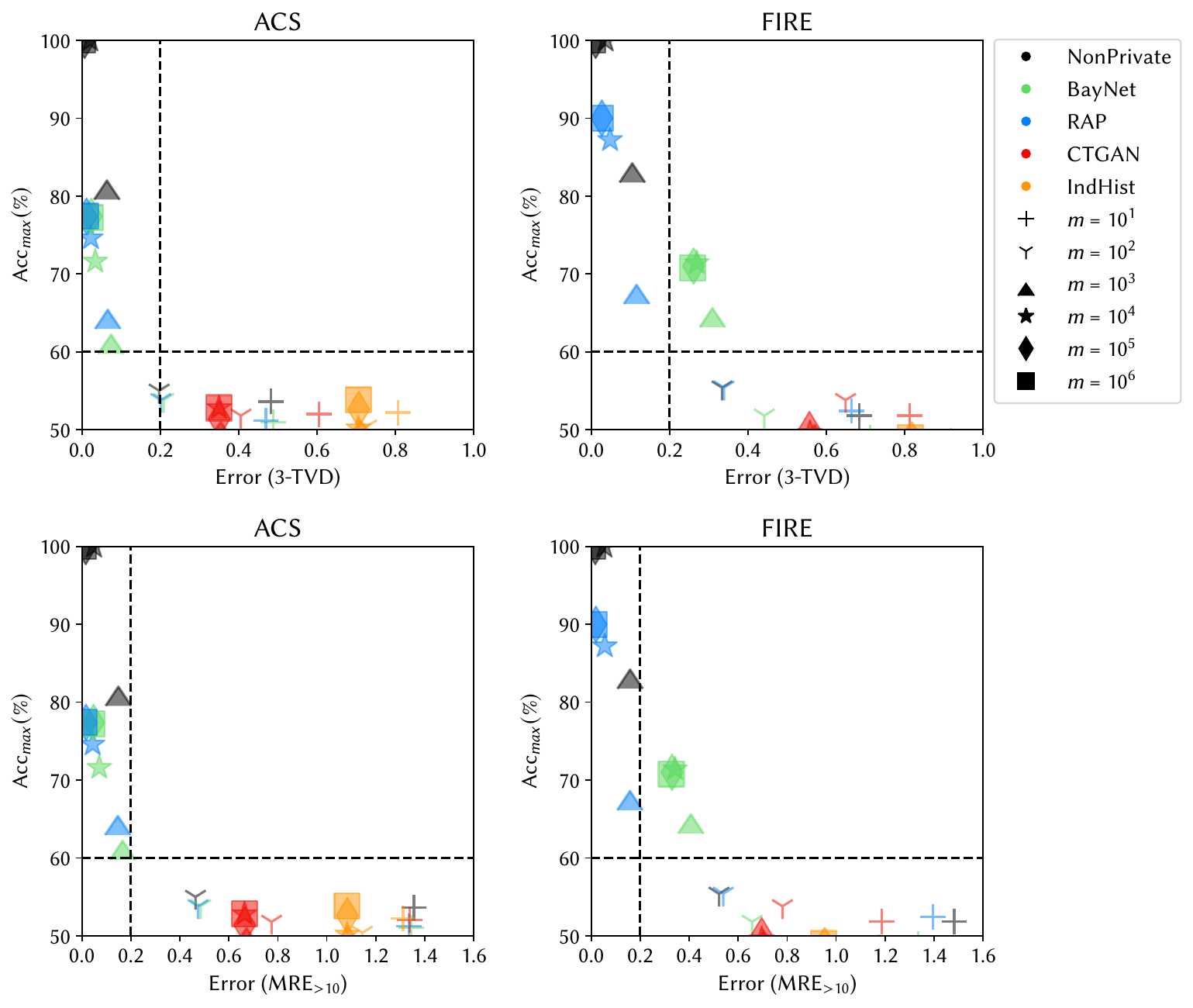}
    \caption{Tradeoff between privacy and utility for $\Advw$}
    \centering
    \label{fig:priv_utility_tradeoff_dcr}
    \end{figure}
    
    \begin{figure}[H]
    \includegraphics[width=\linewidth]{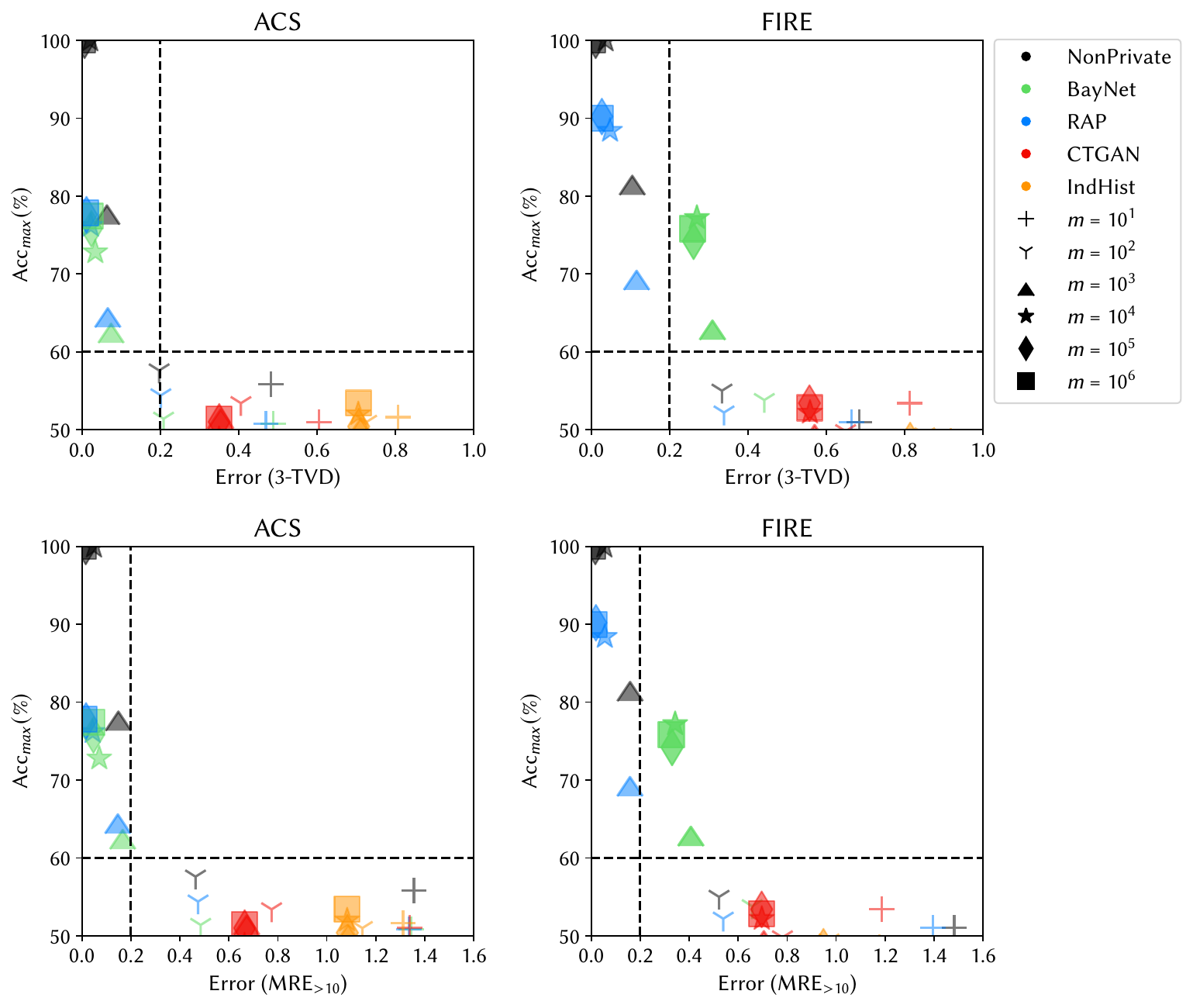}
    \caption{Tradeoff between privacy and utility for $\Advi$}
    \centering
    \label{fig:priv_utility_tradeoff_inference}
    \end{figure}

    \section{All error values for $\Em$ and $\Er$ on synthetic data generation algorithms}
    \label{appsec:exact_error}
    \begin{table}[H]
        \begin{center}
            \setlength\tabcolsep{2pt} %
            \begin{tabular}{c|ccccc}
                \toprule
                $m$ & NonPrivate & BayNet & RAP & CTGAN & IndHist \\
                \midrule
                $10^1$ & $0.48{\pm}0.04$ & $0.49{\pm}0.03$ & $0.47{\pm}0.02$ & $0.61{\pm}0.03$ & $0.81{\pm}0.06$ \\
                $10^2$ & $0.20{\pm}0.02$ & $0.21{\pm}0.01$ & $0.20{\pm}0.01$ & $0.41{\pm}0.03$ & $0.73{\pm}0.06$ \\
                $10^3$ & $0.06{\pm}0.01$ & $0.07{\pm}0.01$ & $0.07{\pm}0.01$ & $0.35{\pm}0.03$ & $0.71{\pm}0.05$ \\
                $10^4$ & $0.02{\pm}0.01$ & $0.03{\pm}0.01$ & $0.02{\pm}0.01$ & $0.35{\pm}0.02$ & $0.71{\pm}0.05$ \\
                $10^5$ & $0.01{\pm}0.01$ & $0.02{\pm}0.01$ & $0.01{\pm}0.01$ & $0.35{\pm}0.02$ & $0.71{\pm}0.05$ \\
                $10^6$ & $0.00{\pm}0.01$ & $0.02{\pm}0.01$ & $0.01{\pm}0.01$ & $0.35{\pm}0.02$ & $0.71{\pm}0.05$ \\
                \bottomrule
            \end{tabular}
            \vspace{0.5cm}
        \end{center}
        \caption{$\Em$ (mean $\pm$ s.d.) for random 3-way marginal queries for ACS dataset}
        \label{table:err_tvd_acs}
    \end{table}
    
    \begin{table}[H]
        \begin{center}
            \setlength\tabcolsep{2pt} %
            \begin{tabular}{c|ccccc}
                \toprule
                $m$ & NonPrivate & BayNet & RAP & CTGAN & IndHist \\
                \midrule
                $10^1$ & $0.68{\pm}0.04$ & $0.71{\pm}0.03$ & $0.66{\pm}0.03$ & $0.81{\pm}0.05$ & $0.92{\pm}0.03$ \\
                $10^2$ & $0.33{\pm}0.02$ & $0.44{\pm}0.01$ & $0.34{\pm}0.02$ & $0.65{\pm}0.04$ & $0.85{\pm}0.04$ \\
                $10^3$ & $0.10{\pm}0.01$ & $0.31{\pm}0.01$ & $0.12{\pm}0.01$ & $0.57{\pm}0.04$ & $0.82{\pm}0.03$ \\
                $10^4$ & $0.04{\pm}0.01$ & $0.27{\pm}0.01$ & $0.05{\pm}0.01$ & $0.56{\pm}0.04$ & $0.81{\pm}0.04$ \\
                $10^5$ & $0.01{\pm}0.01$ & $0.26{\pm}0.01$ & $0.03{\pm}0.01$ & $0.56{\pm}0.04$ & $0.81{\pm}0.04$ \\
                $10^6$ & $0.00{\pm}0.01$ & $0.26{\pm}0.01$ & $0.02{\pm}0.01$ & $0.56{\pm}0.04$ & $0.81{\pm}0.04$ \\
                \bottomrule
            \end{tabular}
            \vspace{0.5cm}
        \end{center}
        \caption{$\Em$ (mean $\pm$ s.d.) for random 3-way marginal queries for FIRE dataset}
        \label{table:err_tvd_fire}
    \end{table}
    
    \begin{table}[H]
        \begin{center}
            \setlength\tabcolsep{2pt} %
            \begin{tabular}{c|ccccc}
                \toprule
                $m$ & NonPrivate & BayNet & RAP & CTGAN & IndHist \\
                \midrule
                $10^1$ & $1.36{\pm}0.08$ & $1.35{\pm}0.06$ & $1.33{\pm}0.05$ & $1.34{\pm}0.07$ & $1.31{\pm}0.10$ \\
                $10^2$ & $0.46{\pm}0.03$ & $0.48{\pm}0.03$ & $0.47{\pm}0.03$ & $0.77{\pm}0.05$ & $1.15{\pm}0.07$ \\
                $10^3$ & $0.15{\pm}0.01$ & $0.17{\pm}0.01$ & $0.15{\pm}0.01$ & $0.67{\pm}0.05$ & $1.09{\pm}0.07$ \\
                $10^4$ & $0.05{\pm}0.01$ & $0.07{\pm}0.01$ & $0.04{\pm}0.01$ & $0.66{\pm}0.05$ & $1.08{\pm}0.07$ \\
                $10^5$ & $0.02{\pm}0.01$ & $0.05{\pm}0.01$ & $0.02{\pm}0.01$ & $0.66{\pm}0.05$ & $1.08{\pm}0.07$ \\
                $10^6$ & $0.00{\pm}0.01$ & $0.04{\pm}0.01$ & $0.01{\pm}0.01$ & $0.66{\pm}0.05$ & $1.08{\pm}0.07$ \\
                \bottomrule
            \end{tabular}
            \vspace{0.5cm}
        \end{center}
        \caption{$\Er$ (mean $\pm$ s.d.) for random 3-way marginal queries for ACS dataset}
        \label{table:err_rel_acs}
    \end{table}
    
    \begin{table}[H]
        \begin{center}
            \setlength\tabcolsep{2pt} %
            \begin{tabular}{c|ccccc}
                \toprule
                $m$ & NonPrivate & BayNet & RAP & CTGAN & IndHist \\
                \midrule
                $10^1$ & $1.48{\pm}0.11$ & $1.34{\pm}0.07$ & $1.39{\pm}0.09$ & $1.19{\pm}0.08$ & $1.18{\pm}0.09$ \\
                $10^2$ & $0.52{\pm}0.05$ & $0.66{\pm}0.04$ & $0.54{\pm}0.05$ & $0.78{\pm}0.05$ & $0.97{\pm}0.05$ \\
                $10^3$ & $0.16{\pm}0.02$ & $0.41{\pm}0.02$ & $0.16{\pm}0.02$ & $0.70{\pm}0.05$ & $0.95{\pm}0.05$ \\
                $10^4$ & $0.05{\pm}0.01$ & $0.34{\pm}0.02$ & $0.06{\pm}0.01$ & $0.70{\pm}0.05$ & $0.95{\pm}0.05$ \\
                $10^5$ & $0.02{\pm}0.01$ & $0.33{\pm}0.02$ & $0.02{\pm}0.01$ & $0.70{\pm}0.05$ & $0.95{\pm}0.05$ \\
                $10^6$ & $0.01{\pm}0.01$ & $0.33{\pm}0.02$ & $0.01{\pm}0.01$ & $0.70{\pm}0.05$ & $0.95{\pm}0.05$ \\
                \bottomrule
            \end{tabular}
            \vspace{0.5cm}
        \end{center}
        \caption{$\Er$ (mean $\pm$ s.d.) for random 3-way marginal queries for FIRE dataset}
        \label{table:err_rel_fire}
    \end{table}

    \section{All error values for $\Em$ and $\Er$ on differentially private synthetic data generation algorithms ($\RAPDP$)}
    \label{appsec:exact_error_dp_rapdp}
    \begin{table}[H]
        \begin{center}
            \setlength\tabcolsep{2pt} %
            \begin{tabular}{c|cccc}
                \toprule
                $m$ & $\varepsilon = 1$ & $\varepsilon = 10$ & $\varepsilon = 100$ \\
                \midrule
                $10^1$ & $0.77{\pm}0.06$ & $0.53{\pm}0.03$ & $0.52{\pm}0.03$ \\
                $10^2$ & $0.66{\pm}0.04$ & $0.28{\pm}0.02$ & $0.23{\pm}0.01$ \\
                $10^3$ & $0.61{\pm}0.04$ & $0.23{\pm}0.01$ & $0.12{\pm}0.01$ \\
                $10^4$ & $0.60{\pm}0.03$ & $0.22{\pm}0.01$ & $0.09{\pm}0.01$ \\
                $10^5$ & $0.60{\pm}0.03$ & $0.22{\pm}0.01$ & $0.09{\pm}0.01$ \\
                $10^6$ & $0.60{\pm}0.03$ & $0.22{\pm}0.01$ & $0.09{\pm}0.01$ \\
                \bottomrule
            \end{tabular}
            \vspace{0.5cm}
        \end{center}
        \caption{$\Em$ (mean $\pm$ s.d.) for random 3-way marginal queries for ACS dataset}
    \end{table}
    
    \begin{table}[H]
        \begin{center}
            \setlength\tabcolsep{2pt} %
            \begin{tabular}{c|cccc}
                \toprule
                $m$ & $\varepsilon = 1$ & $\varepsilon = 10$ & $\varepsilon = 100$ \\
                \midrule
                $10^1$ & $0.97{\pm}0.04$ & $0.77{\pm}0.04$ & $0.71{\pm}0.03$ \\
                $10^2$ & $0.93{\pm}0.03$ & $0.57{\pm}0.03$ & $0.43{\pm}0.02$ \\
                $10^3$ & $0.88{\pm}0.03$ & $0.48{\pm}0.02$ & $0.29{\pm}0.01$ \\
                $10^4$ & $0.87{\pm}0.03$ & $0.46{\pm}0.02$ & $0.27{\pm}0.01$ \\
                $10^5$ & $0.87{\pm}0.03$ & $0.46{\pm}0.02$ & $0.27{\pm}0.01$ \\
                $10^6$ & $0.87{\pm}0.03$ & $0.46{\pm}0.02$ & $0.27{\pm}0.01$ \\
                \bottomrule
            \end{tabular}
            \vspace{0.5cm}
        \end{center}
        \caption{$\Em$ (mean $\pm$ s.d.) for random 3-way marginal queries for FIRE dataset}
    \end{table}
    
    \begin{table}[H]
        \begin{center}
            \setlength\tabcolsep{2pt} %
            \begin{tabular}{c|cccc}
                \toprule
                $m$ & $\varepsilon = 1$ & $\varepsilon = 10$ & $\varepsilon = 100$ \\
                \midrule
                $10^1$ & $1.24{\pm}0.06$ & $1.23{\pm}0.06$ & $1.33{\pm}0.05$ \\
                $10^2$ & $0.84{\pm}0.03$ & $0.55{\pm}0.03$ & $0.48{\pm}0.03$ \\
                $10^3$ & $0.76{\pm}0.03$ & $0.39{\pm}0.02$ & $0.21{\pm}0.01$ \\
                $10^4$ & $0.75{\pm}0.03$ & $0.37{\pm}0.02$ & $0.16{\pm}0.01$ \\
                $10^5$ & $0.75{\pm}0.03$ & $0.37{\pm}0.02$ & $0.15{\pm}0.01$ \\
                $10^6$ & $0.75{\pm}0.03$ & $0.37{\pm}0.02$ & $0.15{\pm}0.01$ \\
                \bottomrule
            \end{tabular}
            \vspace{0.5cm}
        \end{center}
        \caption{$\Er$ (mean $\pm$ s.d.) for random 3-way marginal queries for ACS dataset}
    \end{table}
    
    \begin{table}[H]
        \begin{center}
            \setlength\tabcolsep{2pt} %
            \begin{tabular}{c|cccc}
                \toprule
                $m$ & $\varepsilon = 1$ & $\varepsilon = 10$ & $\varepsilon = 100$ \\
                \midrule
                $10^1$ & $1.02{\pm}0.03$ & $1.18{\pm}0.08$ & $1.27{\pm}0.09$ \\
                $10^2$ & $0.96{\pm}0.02$ & $0.71{\pm}0.03$ & $0.57{\pm}0.04$ \\
                $10^3$ & $0.95{\pm}0.02$ & $0.63{\pm}0.03$ & $0.35{\pm}0.02$ \\
                $10^4$ & $0.95{\pm}0.02$ & $0.61{\pm}0.03$ & $0.32{\pm}0.02$ \\
                $10^5$ & $0.95{\pm}0.02$ & $0.61{\pm}0.03$ & $0.31{\pm}0.02$ \\
                $10^6$ & $0.95{\pm}0.02$ & $0.61{\pm}0.03$ & $0.31{\pm}0.02$ \\
                \bottomrule
            \end{tabular}
            \vspace{0.5cm}
        \end{center}
        \caption{$\Er$ (mean $\pm$ s.d.) for random 3-way marginal queries for FIRE dataset}
    \end{table}

    \section{All error values for $\Em$ and $\Er$ on differentially private synthetic data generation algorithms (PrivBayes)}
    \label{appsec:exact_error_dp_privbayes}
    \begin{table}[H]
        \begin{center}
            \setlength\tabcolsep{2pt} %
            \begin{tabular}{c|cccc}
                \toprule
                $m$ & $\varepsilon = 1$ & $\varepsilon = 10$ & $\varepsilon = 100$ \\
                \midrule
                $10^1$ & $0.85{\pm}0.04$ & $0.75{\pm}0.05$ & $0.67{\pm}0.05$ \\
                $10^2$ & $0.72{\pm}0.02$ & $0.59{\pm}0.04$ & $0.41{\pm}0.03$ \\
                $10^3$ & $0.68{\pm}0.02$ & $0.55{\pm}0.04$ & $0.34{\pm}0.03$ \\
                $10^4$ & $0.68{\pm}0.02$ & $0.54{\pm}0.04$ & $0.33{\pm}0.03$ \\
                $10^5$ & $0.67{\pm}0.02$ & $0.54{\pm}0.04$ & $0.33{\pm}0.03$ \\
                $10^6$ & $0.67{\pm}0.02$ & $0.54{\pm}0.04$ & $0.33{\pm}0.03$ \\
                \bottomrule
            \end{tabular}
            \vspace{0.5cm}
        \end{center}
        \caption{$\Em$ (mean $\pm$ s.d.) for random 3-way marginal queries for ACS dataset}
    \end{table}
    
    \begin{table}[H]
        \begin{center}
            \setlength\tabcolsep{2pt} %
            \begin{tabular}{c|cccc}
                \toprule
                $m$ & $\varepsilon = 1$ & $\varepsilon = 10$ & $\varepsilon = 100$ \\
                \midrule
                $10^1$ & $0.99{\pm}0.02$ & $0.91{\pm}0.04$ & $0.72{\pm}0.03$ \\
                $10^2$ & $0.95{\pm}0.01$ & $0.82{\pm}0.03$ & $0.58{\pm}0.03$ \\
                $10^3$ & $0.90{\pm}0.02$ & $0.76{\pm}0.02$ & $0.48{\pm}0.03$ \\
                $10^4$ & $0.89{\pm}0.02$ & $0.75{\pm}0.02$ & $0.45{\pm}0.03$ \\
                $10^5$ & $0.89{\pm}0.01$ & $0.75{\pm}0.02$ & $0.45{\pm}0.03$ \\
                $10^6$ & $0.89{\pm}0.01$ & $0.75{\pm}0.02$ & $0.45{\pm}0.03$ \\
                \bottomrule
            \end{tabular}
            \vspace{0.5cm}
        \end{center}
        \caption{$\Em$ (mean $\pm$ s.d.) for random 3-way marginal queries for FIRE dataset}
    \end{table}
    
    \begin{table}[H]
        \begin{center}
            \setlength\tabcolsep{2pt} %
            \begin{tabular}{c|cccc}
                \toprule
                $m$ & $\varepsilon = 1$ & $\varepsilon = 10$ & $\varepsilon = 100$ \\
                \midrule
                $10^1$ & $1.24{\pm}0.04$ & $1.27{\pm}0.06$ & $1.27{\pm}0.06$ \\
                $10^2$ & $0.85{\pm}0.03$ & $0.75{\pm}0.04$ & $0.58{\pm}0.03$ \\
                $10^3$ & $0.77{\pm}0.02$ & $0.64{\pm}0.04$ & $0.39{\pm}0.04$ \\
                $10^4$ & $0.77{\pm}0.02$ & $0.62{\pm}0.04$ & $0.37{\pm}0.03$ \\
                $10^5$ & $0.77{\pm}0.02$ & $0.62{\pm}0.04$ & $0.37{\pm}0.03$ \\
                $10^6$ & $0.77{\pm}0.02$ & $0.62{\pm}0.04$ & $0.37{\pm}0.03$ \\
                \bottomrule
            \end{tabular}
            \vspace{0.5cm}
        \end{center}
        \caption{$\Er$ (mean $\pm$ s.d.) for random 3-way marginal queries for ACS dataset}
    \end{table}
    
    \begin{table}[H]
        \begin{center}
            \setlength\tabcolsep{2pt} %
            \begin{tabular}{c|cccc}
                \toprule
                $m$ & $\varepsilon = 1$ & $\varepsilon = 10$ & $\varepsilon = 100$ \\
                \midrule
                $10^1$ & $1.03{\pm}0.03$ & $1.06{\pm}0.05$ & $1.23{\pm}0.08$ \\
                $10^2$ & $0.94{\pm}0.02$ & $0.83{\pm}0.03$ & $0.66{\pm}0.03$ \\
                $10^3$ & $0.94{\pm}0.02$ & $0.81{\pm}0.03$ & $0.51{\pm}0.04$ \\
                $10^4$ & $0.94{\pm}0.02$ & $0.81{\pm}0.03$ & $0.49{\pm}0.04$ \\
                $10^5$ & $0.94{\pm}0.02$ & $0.81{\pm}0.03$ & $0.49{\pm}0.04$ \\
                $10^6$ & $0.94{\pm}0.02$ & $0.81{\pm}0.03$ & $0.49{\pm}0.04$ \\
                \bottomrule
            \end{tabular}
            \vspace{0.5cm}
        \end{center}
        \caption{$\Er$ (mean $\pm$ s.d.) for random 3-way marginal queries for FIRE dataset}
    \end{table}
\fi

\end{document}